\pdfoutput=1

\PassOptionsToPackage{table,xcdraw}{xcolor}

\documentclass[11pt]{article}
\usepackage{booktabs}
\usepackage{multirow}
\usepackage{makecell}

\usepackage{acl}
\usepackage{lscape}

\usepackage{times}
\usepackage{latexsym}

\usepackage[T1]{fontenc}

\usepackage[utf8]{inputenc}

\usepackage{microtype}

\usepackage{inconsolata}

\usepackage{mathtools}
\usepackage{amssymb}
\usepackage{amsmath}
\usepackage{mdframed}
\usepackage{fdsymbol} 
\usepackage{pifont} 
\usepackage{multirow} 
\usepackage{comment}
\usepackage[most]{tcolorbox} 
\usepackage{longtable}
\usepackage[normalem]{ulem}
\useunder{\uline}{\ul}{}
\usepackage{colortbl} 
\usepackage{tablefootnote}
\usepackage{graphicx} 
\usepackage[table]{xcolor} 
\usepackage{xcolor,amsmath}
\usepackage[ruled]{algorithm2e}
\usepackage{colortbl}
\usepackage[group-separator={,}]{siunitx}
\usepackage{comment}
\usepackage{arydshln} 

\definecolor{darkgreen}{rgb}{0.0, 0.6, 0.0}

\definecolor{customRed}{RGB}{230,62,48}    
\definecolor{customBlue}{RGB}{2,129,197}   
\definecolor{customGreen}{RGB}{112,173,71}  
\definecolor{customOrange}{RGB}{255,179,26} 

\definecolor{customPurple}{RGB}{128, 0, 128}    
\definecolor{customYellow}{RGB}{255, 205, 0}    
\definecolor{customCyan}{RGB}{0, 139, 139}      
\definecolor{customMagenta}{RGB}{255, 0, 255}   
\definecolor{customGray}{RGB}{128, 128, 128}    
\definecolor{customNavy}{RGB}{0, 0, 180}        
\definecolor{customTeal}{RGB}{0, 128, 128}      
\definecolor{customOlive}{RGB}{128, 128, 0}     
\definecolor{customMaroon}{RGB}{128, 0, 0}      
\definecolor{customAqua}{RGB}{0, 255, 255}      

\newcommand{\secref}[1]{\S\ref{#1}}

\newmdenv[
  backgroundcolor=red!05,
  linecolor=quoteborder,
  skipabove=1em,
  skipbelow=0em,
  leftline=true,
  topline=false,
  bottomline=false,
  rightline=false,
  linecolor=red!66,
  linewidth=4pt
]{githubquote}

\newcommand{\cmark}{\textcolor{darkgreen}{\ding{51}}} 
\newcommand{\xmark}{\textcolor{customRed}{\ding{55}}} 

\newcommand{\ft}{\textbf{F$_1$@3}\xspace}
\newcommand{\ff}{\textbf{F$_1$@5}\xspace}
\newcommand{\fm}{\textbf{F$_1$@$\mathcal{M}$}\xspace}
\newcommand{\fo}{\textbf{F$_1$@$\mathcal{O}$}\xspace}

\newcommand{\ThickVLine}{\rule{0.65pt}{2.25ex}} 

\SetKwComment{Comment}{\color{black!50!white}// }{}
\newcommand{\assign}{\leftarrow}
\newcommand{\var}{\texttt}
\newcommand{\FuncCall}[2]{\texttt{\bfseries #1(#2)}}
\SetKwProg{Function}{function}{}{}

\title{\includegraphics[width=0.60cm, height=0.65cm]{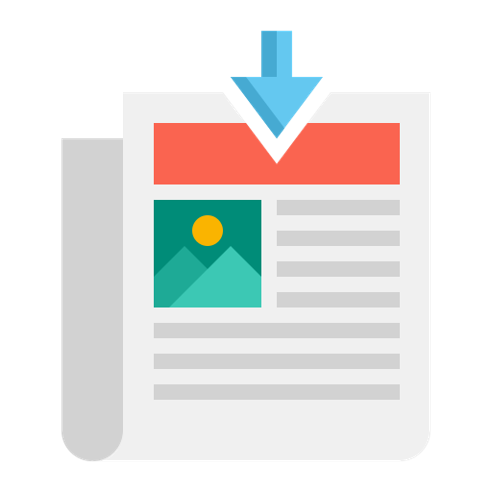} \textcolor{customNavy}{XL}-\textcolor{darkgreen}{Head}\textcolor{customBlue}{Tags}: Leveraging Multimodal Retrieval Augmentation for the Multilingual Generation of News Headlines and Tags}

\author{
Faisal Tareque Shohan$^{\clubsuit}$\thanks{\ \textbf{Equal contribution.}} \enspace Mir Tafseer Nayeem$^{\spadesuit}$\footnotemark[1]\thanks{\ Corresponding author.} \\ 
\bf Samsul Islam$^{\clubsuit}$ \enspace Abu Ubaida Akash$^{\diamondsuit}$ \enspace Shafiq Joty$^{\varheartsuit\vardiamondsuit}$\\
$^\clubsuit$Ahsanullah University of Science \& Technology \enspace 
$^\spadesuit$University of Alberta\\
$^\diamondsuit$Université de Sherbrooke \enspace
$^\varheartsuit$Salesforce Research \enspace $^\vardiamondsuit$Nanyang Technological University\\
\texttt{faisaltareque@hotmail.com} \enspace \texttt{mnayeem@ualberta.ca} \enspace  \texttt{samsulratul98@gmail.com} \\ \texttt{abu.ubaida.akash@usherbrooke.ca} \enspace \texttt{ sjoty@salesforce.com} \\
}

\begin{document}
\maketitle
\begin{abstract}

Millions of news articles published online daily can overwhelm readers. Headlines and entity (topic) tags are essential for guiding readers to decide if the content is worth their time. While headline generation has been extensively studied, tag generation remains largely unexplored, yet it offers readers better access to topics of interest. The need for conciseness in capturing readers' attention necessitates improved content selection strategies for identifying salient and relevant segments within lengthy articles, thereby guiding language models effectively. To address this, we propose to leverage auxiliary information such as images and captions embedded in the articles to retrieve relevant sentences and utilize instruction tuning with variations to generate both headlines and tags for news articles in a multilingual context. To make use of the auxiliary information, we have compiled a dataset named \texttt{\textbf{XL-HeadTags}}, which includes \num{20} languages across \num{6} diverse language families. Through extensive evaluation, we demonstrate the effectiveness of our \emph{plug-and-play} multimodal-multilingual retrievers for both tasks. Additionally, we have developed a suite of tools for processing and evaluating multilingual texts, significantly contributing to the research community by enabling more accurate and efficient analysis across languages.\footnote{Our code, dataset, model checkpoints, developed tools are available at \href{https://github.com/faisaltareque/XL-HeadTags}{XL-HeadTags}.
}
\end{abstract}

\section{Introduction}

The headline serves as a concise and attention-grabbing summary of a news article. Articles with compelling headlines are more likely to attract increased views or shares \cite{10.1145/3366423.3380247, Song_Shuai_Yeh_Wu_Ku_Peng_2020_Attractive_or_Faithful}.  Unlike  summaries, which provide a broad overview \cite{nayeem-etal-2018-abstractive}, headlines aim to produce brief and engaging statements that draw readers into the full article \cite{xu-etal-2019-clickbait, zhang-yang-2023-mediahg, akash-etal-2023-shironaam}. They are often the most-read part of the article, guiding readers in deciding whether the content merits their time \cite{Transformer-Based-Models-for-News-Headline-Generation}.

Another important feature of a news article are 
topic tags or \texttt{``semantic markers''}, which serve as dynamic connectors and navigational aids, enhancing the coherence and accessibility of information across articles. The task of tag generation is related to keyphrase generation \cite{meng-etal-2017-deep}. While keyphrases summarize the main themes succinctly of an article, tags provide a broader overview, guiding readers to related articles and facilitating navigation through connected themes. Despite their significant role, the generation of news article tags has remained unexplored in existing literature.

\begin{figure}[t]
    \centering
    \includegraphics[scale = 0.40]{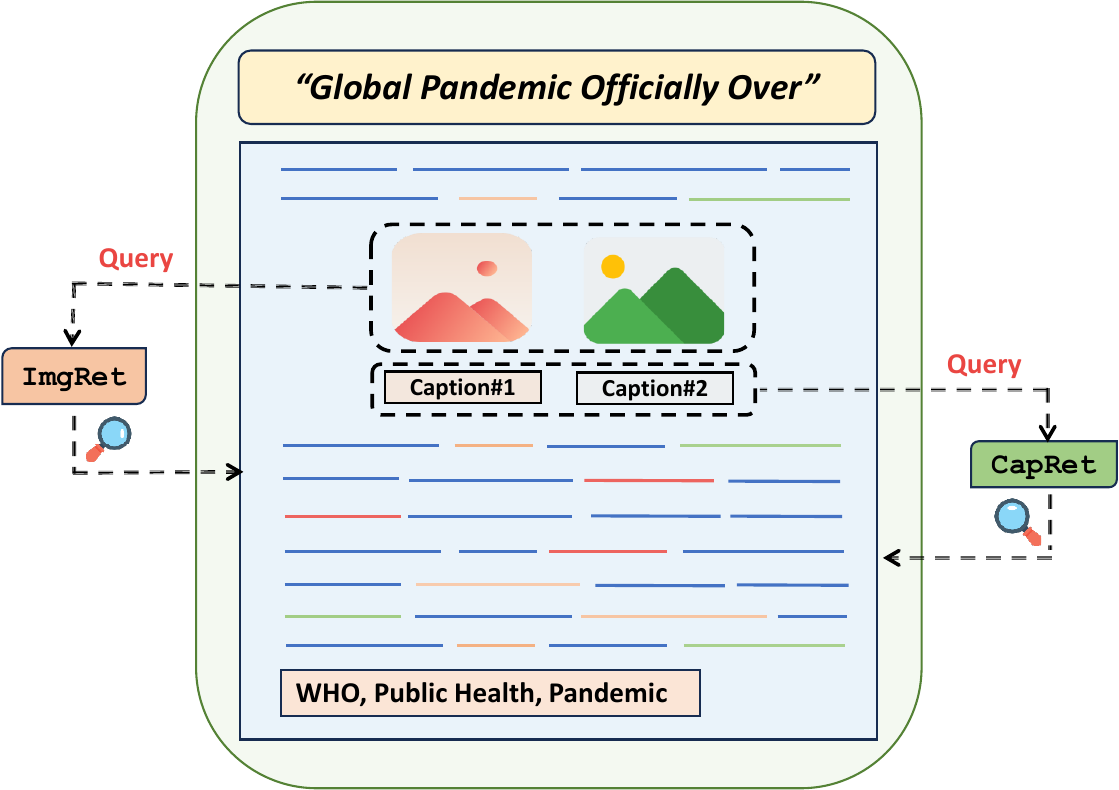}
    \caption{Our content selection approach. Auxiliary information, including images and captions, is used as \emph{queries} to extract salient and relevant sentences from documents via two modules: \texttt{\textbf{ImgRet}} for images (\texttt{visual modality}) and \texttt{\textbf{CapRet}} for image captions (\texttt{textual modality}). Our modules are designed as \texttt{plug-and-play} components that can be integrated with language models of any size and type.}
    \label{fig:idea-illustration}
\end{figure}

In this work, we introduce a unified framework for generating headlines and tags for news articles in a multimodal-multilingual context, covering \num{20} languages across \num{6} diverse language families. Despite the emergence of open-source Large Language Models (LLMs) like \texttt{Llama} \cite{touvron2023llama} and \texttt{Mistral} \cite{jiang2023mistral}, multilingual models such as \texttt{mT5} \cite{xue-etal-2021-mt5} and \texttt{mT0} \cite{mt0-muennighoff2022crosslingual}, fine-tuned with task-specific data, remain the preferred solution for multilingual tasks, especially for  low-resource languages \cite{ahuja-etal-2023-mega, zhao2024llama, aggarwal2024maple}. News articles often include extensive content, incorporating additional details, author quotes, historical context, and advertisements, among others. This poses a challenge in identifying article segments that are both salient and relevant for the creation of highly concise outputs like headlines and tags.  This complication leads to what is known  as the \texttt{``Lost-in-the-Context''} problem \cite{liu2023lost}, wherein vital information embedded within lengthy documents is frequently overlooked by the models \cite{ravaut2024context}. 

Fortunately, it is now very common for both online and printed news media to use multimedia content to enhance visibility, support, and context for articles \cite{oostdijk-etal-2020-connection}. Digital assets, like images, often serve as thumbnails across social media, blogs, and various platforms. Equally important are the captions accompanying these images, which not only clarify and enrich the image but also optimize articles for search engines and make news more accessible to those with vision impairments \cite{liu-etal-2021-visual}. This has motivated us to utilize auxiliary information (images and captions {as illustrated in Figure~\ref{fig:idea-illustration}}) to distill salient and localized information from news articles in a multimodal-multilingual context using the \texttt{\textbf{CLIP-ViT-B32}} model \cite{pmlr-v139-radford21a}. This model aligns text and images within a unified dense vector space. Our method is inspired by the \texttt{\textbf{Cognitive Load Theory}} \cite{SWELLER201137}, reflecting how humans employ visual cues and summaries to understand the essence of lengthy texts without being overwhelmed by information. Building on this, with the contents selected using our multimodal retrievers,  we utilize instruction tuning to generate both headlines and tags for news articles  in a multilingual context.


Our contributions are summarized as follows:

\begin{itemize}
    \item We compile the \texttt{\textbf{XL-HeadTags}} dataset for headline and tags generation tasks, expanding it to include \num{20} languages across \num{6} diverse language families (\secref{sec:dataset}).

    \item We present a new approach to content selection that utilizes auxiliary information from both textual and visual modalities to identify the most salient content within news articles in a multilingual setting. Our modules are crafted as \texttt{plug-and-play} components, allowing for seamless integration with language models of any size and type (\secref{sec:task-definition-MultiRAGen}).

    \item We utilize instruction tuning to generate both headline and tag words. Our model is capable of producing tag words in both controlled and unrestricted manners through instructions (\secref{sec:instruction-tuning}). Furthermore, we introduce novel tag words evaluation metrics designed to evaluate scenarios of both controlled and unrestricted generation effectively (\secref{sec:proposed-tag-eval}).

    \item We also develop tools by accumulating open-source resources for processing and evaluating multilingual texts, making it an easy-to-use one-stop destination. These tools include \textbf{(1)} Multilingual ROUGE Scorer, \textbf{(2)} Multilingual Sentence Tokenizer, and \textbf{(3)} Multilingual Stemmer. These resources are invaluable to the research community focusing on multilingual NLP (\secref{sec:tools}).
\end{itemize}

\section{XL-HeadTags: Dataset \& Tasks}
\label{sec:dataset-tasks}

\begin{figure*}[t]
    \centering
    \includegraphics[scale = 0.45]{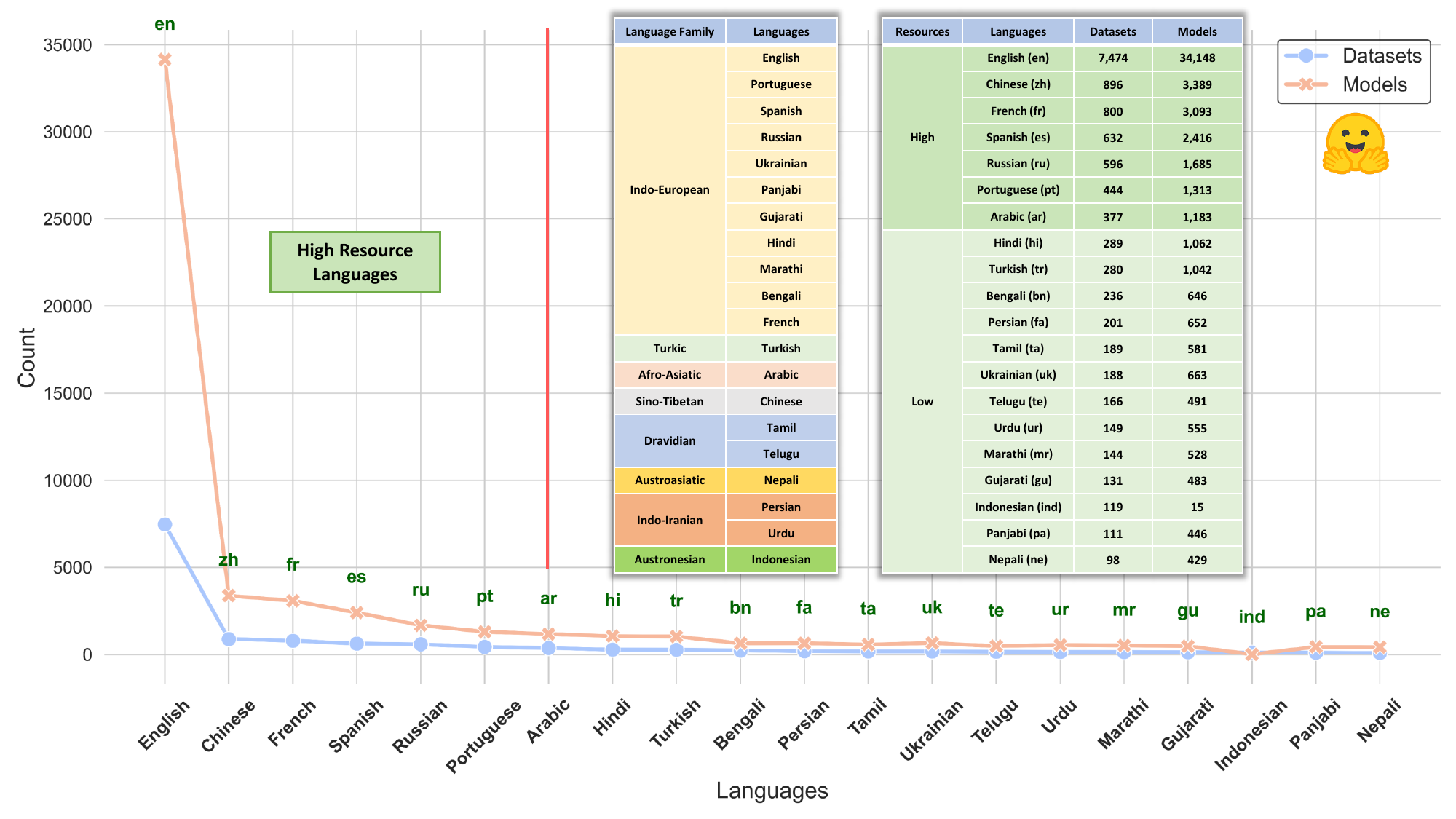}
    \caption{Distribution of Datasets and Models Across Languages. Data were sourced from the \textbf{Huggingface} resource ranking (\url{https://huggingface.co/languages}) as of February 5, 2024.}
    \label{fig:huggingface-resources}
\end{figure*}

\subsection{Dataset}
\label{sec:dataset}

Our study focuses on exploring techniques for multilingual and multimodal retrieval, aiming to extract salient and localized information from documents. This is intended to support the generation of succinct headlines and tags. Our goal is to utilize auxiliary information, such as images and image captions, as queries to retrieve salient information from the document. To achieve our goals, we explore existing large-scale multilingual and multimodal datasets to identify auxiliary information, such as images and their captions. Table~\ref{table:dataset-comparison} highlights several well-known datasets that serve summarization purposes. We choose \texttt{M3LS} \cite{m3ls-verma-etal-2023-large} and \texttt{XL-Sum} \cite{hasan-etal-2021-xl} as our primary data sources. Since both datasets share the BBC as their source, we anticipate minimal distributional and structural shifts, which could enhance the coherence and efficiency of our retrieval process.

\texttt{M3LS} is a large-scale dataset designed for Multimodal Multilingual Summarization, featuring headlines, articles, summaries, images, captions and tag words. In contrast, \texttt{XL-Sum} focuses on Multilingual Abstractive Summarization, comprising headlines, articles, summaries, and news links. It is noteworthy that although the \texttt{M3LS} dataset includes images, captions, and tag words, these were not leveraged for the development of their models. Instead of recrawling the data points, we specifically utilized the auxiliary information from \texttt{M3LS}, for our retrieval augmentation framework. Furthermore, to expand our dataset to additional language families, we incorporated Arabic, Turkish, and Persian from the 
\texttt{XL-Sum} dataset. A notable limitation of \texttt{XL-Sum} is its absence of images, image captions, and tag words. To address this gap, we utilized the news URLs provided in the \texttt{XL-Sum} dataset to gather the necessary auxiliary information for our framework. We employ a \texttt{scrapy}\footnote{\url{https://scrapy.org}} framework-based web crawler to systematically collect detailed information, including images, captions, and tag words from news articles. Our criteria require the presence of key elements (headlines, articles, images, captions, and tags) in each data point.

\begin{table}[t]
\centering
\resizebox{0.9\linewidth}{!} 
{ 
\centering
\renewcommand{\arraystretch}{1} 
\begin{tabular}{c|ccccccccc}
\Xhline{4\arrayrulewidth}
\multicolumn{1}{c}{}                                    &                                                                                     &                                                                                   &                                  & \multicolumn{6}{c}{\textbf{Auxiliary Information}}                                                                                                                   \\ \cline{5-10} 
\multicolumn{1}{c}{}                                    &                                                                                     &                                                                                   &                                  & \multicolumn{2}{c}{\textbf{Img}}                                & \multicolumn{2}{c}{\textbf{Cap}}                                & \multicolumn{2}{c}{\textbf{Tag}} \\ \cline{5-10} 
\multicolumn{1}{c}{\multirow{-3}{*}{\textbf{Datasets}}} & \multirow{-3}{*}{\textbf{\begin{tabular}[c]{@{}c@{}}Multi-\\ lingual\end{tabular}}} & \multirow{-3}{*}{\textbf{\begin{tabular}[c]{@{}c@{}}Multi-\\ modal\end{tabular}}} & \multirow{-3}{*}{\textbf{Task}}  & \textbf{A} & \multicolumn{1}{c|}{\textbf{U}}                    & \textbf{A} & \multicolumn{1}{c|}{\textbf{U}}                    & \textbf{A}      & \textbf{U}     \\ \Xhline{4\arrayrulewidth}
\rowcolor[HTML]{DCDCDC} 
\cellcolor[HTML]{DCDCDC} \textbf{\texttt{MSMO}} \shortcite{zhu-etal-2018-msmo}                            & \cellcolor[HTML]{DCDCDC}\xmark                                                       & \cellcolor[HTML]{DCDCDC}\cmark                                                     & \cellcolor[HTML]{DCDCDC}Summ      & \cmark      & \multicolumn{1}{c|}{\cellcolor[HTML]{DCDCDC}\cmark} & \cmark      & \multicolumn{1}{c|}{\cellcolor[HTML]{DCDCDC}\xmark} & \xmark           & \xmark          \\ 
\rowcolor[HTML]{FFFFFF} 
\textbf{\texttt{E-DM}} \shortcite{chen-zhuge-2018-abstractive}                                                    & \xmark                                                                               & \cmark                                                                             & Summ                              & \cmark      & \multicolumn{1}{c|}{\cellcolor[HTML]{FFFFFF}\cmark} & \cmark      & \multicolumn{1}{c|}{\cellcolor[HTML]{FFFFFF}\cmark} & \xmark           & \xmark          \\ 
\rowcolor[HTML]{DCDCDC} 
\cellcolor[HTML]{DCDCDC} \textbf{\texttt{XSum}} \shortcite{narayan-etal-2018-dont}                            & \xmark                                                                               & \xmark                                                                             & \cellcolor[HTML]{DCDCDC}Summ      & \xmark      & \multicolumn{1}{c|}{\cellcolor[HTML]{DCDCDC}\xmark} & \xmark      & \multicolumn{1}{c|}{\cellcolor[HTML]{DCDCDC}\xmark} & \xmark           & \xmark          \\ 
\rowcolor[HTML]{FFFFFF} 
\textbf{\texttt{CNN/DM}} \shortcite{see-etal-2017-get}                                                   & \xmark                                                                               & \xmark                                                                             & Summ                              & \xmark      & \multicolumn{1}{c|}{\cellcolor[HTML]{FFFFFF}\xmark} & \xmark      & \multicolumn{1}{c|}{\cellcolor[HTML]{FFFFFF}\xmark} & \xmark           & \xmark          \\ 
\rowcolor[HTML]{DCDCDC} 
\cellcolor[HTML]{DCDCDC} \textbf{\texttt{MMSS}} \shortcite{ijcai2018p577}                             & \xmark                                                                               & \cmark                                                                             & \cellcolor[HTML]{DCDCDC}Summ      & \cmark      & \multicolumn{1}{c|}{\cellcolor[HTML]{DCDCDC}\cmark} & \xmark      & \multicolumn{1}{c|}{\cellcolor[HTML]{DCDCDC}\xmark} & \xmark           & \xmark          \\
\rowcolor[HTML]{FFFFFF} 
\cellcolor[HTML]{FFFFFF} \textbf{\texttt{MLASK}} \shortcite{krubinski-pecina-2023-mlask}                             & \xmark                                                                               & \cmark                                                                             & \cellcolor[HTML]{FFFFFF}Summ      & \cmark      & \multicolumn{1}{c|}{\cellcolor[HTML]{FFFFFF}\cmark} & \xmark      & \multicolumn{1}{c|}{\cellcolor[HTML]{FFFFFF}\xmark} & \xmark           & \xmark          \\ 
\rowcolor[HTML]{DCDCDC} 
\textbf{\texttt{MLSUM}} \shortcite{scialom-etal-2020-mlsum}                                                   & \cmark                                                                               & \xmark                                                                             & Summ                              & \xmark      & \multicolumn{1}{c|}{\cellcolor[HTML]{DCDCDC}\xmark} & \xmark      & \multicolumn{1}{c|}{\cellcolor[HTML]{DCDCDC}\xmark} & \cmark          & \xmark          \\ 
\rowcolor[HTML]{FFFFFF} 
\cellcolor[HTML]{FFFFFF} \textbf{\texttt{XL-SUM}} \shortcite{hasan-etal-2021-xl}                          & \cmark                                                                               & \xmark                                                                             & \cellcolor[HTML]{FFFFFF}Summ      & \xmark      & \multicolumn{1}{c|}{\cellcolor[HTML]{FFFFFF}\xmark} & \xmark      & \multicolumn{1}{c|}{\cellcolor[HTML]{FFFFFF}\xmark} & \xmark           & \xmark          \\ 
\hdashline
\rowcolor[HTML]{DCDCDC} 
\textbf{\texttt{M3LS}} \shortcite{m3ls-verma-etal-2023-large}                                                     & \cmark                                                                               & \cmark                                                                             & Summ                              & \cmark      & \multicolumn{1}{c|}{\cellcolor[HTML]{DCDCDC}\xmark} & \cmark      & \multicolumn{1}{c|}{\cellcolor[HTML]{DCDCDC}\xmark} & \cmark           & \xmark          \\ 
\rowcolor[HTML]{FFFFFF} 
\cellcolor[HTML]{FFFFFF} \textbf{\texttt{XL-HeadTags}}                           & \cmark                                                                               & \cmark                                                                             & \cellcolor[HTML]{FFFFFF}Headline \& Tags & \cmark      & \multicolumn{1}{c|}{\cellcolor[HTML]{FFFFFF}\cmark} & \cmark      & \multicolumn{1}{c|}{\cellcolor[HTML]{FFFFFF}\cmark} & \cmark           & \cmark          \\ \Xhline{4\arrayrulewidth}
\end{tabular}
}
\caption{Comparison of the \textbf{\texttt{XL-HeadTags}} (\textit{ours}) dataset with existing multi-lingual and multi-modal datasets. \textbf{`A'} means whether the auxiliary information is \textbf{``Available''} and \textbf{`U'} indicates whether the information is \textbf{``Utilized''} by the models. \textbf{``Summ''} denotes Summarization Task.}
\label{table:dataset-comparison}
\end{table}

Our \textbf{\texttt{XL-HeadTags}} dataset consists of \num{20} languages across six diverse language families. Detailed statistics of our dataset are presented in Table \ref{table:datas-stats-1} and described in the Appendix \ref{sec:dataset-processing}. Additionally, we have developed a mechanism to classify these languages into high-resource and low-resource categories. \citet{conneau-etal-2020-unsupervised} utilized CommonCrawl (CC) pretraining data, quantified in gigabytes (GB), to highlight the disparities between high- and low-resource languages. This differentiation is determined by examining the volume of pretraining data used in developing pretrained language models. In contrast, we define high-resource and low-resource languages based on the availability of task-specific resources. This includes both datasets and pretrained and/or task-specific fine-tuned models, as per the Huggingface resource ranking\footnote{\url{https://huggingface.co/languages}}, which offers real-time and up-to-date information. Languages ranked in the top \num{10} of this list are classified as high-resource, while the others are low-resource. Figure \ref{fig:huggingface-resources} illustrates the distribution of datasets and models across languages.

\begin{table*}[t!]
\centering
\scalebox{.68}{
\begin{tabular}{clccccccccccccc}
\Xhline{4\arrayrulewidth}
                                                                                      & \multicolumn{1}{c}{}                                     &                                                                             &                                                                              &                                                                             &                                                                             & \multicolumn{5}{c}{\textbf{\% of novel n-grams}}                                                                               &                                 &                                                                               &                                                                             &                                                                                    \\ \cline{8-11}
\multirow{-2}{*}{\textbf{\begin{tabular}[c]{@{}c@{}}Language\\  Family\end{tabular}}} & \multicolumn{1}{c}{\multirow{-2}{*}{\textbf{Languages}}} & \multirow{-2}{*}{\textbf{\#Samples}} & \multirow{-2}{*}{\textbf{\begin{tabular}[c]{@{}c@{}}Avg.\\ Word\end{tabular}}} & \multirow{-2}{*}{\textbf{\begin{tabular}[c]{@{}c@{}}Avg.\\ Sent\end{tabular}}} & \multirow{-2}{*}{\textbf{\begin{tabular}[c]{@{}c@{}}Avg.\\ Tok\end{tabular}}} & \multirow{-2}{*}{\textbf{\begin{tabular}[c]{@{}c@{}} Avg. H\\ Word\end{tabular}}} & \textbf{uni}                  & \textbf{bi}                   & \textbf{tri}                  & \textbf{quad}                 & \multirow{-2}{*}{\textbf{CR}} & \multirow{-2}{*}{\textbf{\begin{tabular}[c]{@{}c@{}}Avg.\\ I / C\end{tabular}}} & \multirow{-2}{*}{\textbf{\begin{tabular}[c]{@{}c@{}}Avg.\\ Tag W\end{tabular}}} & \multirow{-2}{*}{\textbf{\begin{tabular}[c]{@{}c@{}}\% Pre\\ Tag W\end{tabular}}} \\ \Xhline{4\arrayrulewidth}
\multicolumn{1}{c|}{}                                                                 & \multicolumn{1}{l|}{English}                             & 200,813 &569                                                                         & 15.34                                                                        & 909                                                                         & 8.43                                                                       & 33.27                         & 84.22                         & 96.54                         & 98.98                         & 98.51                           & 2.85                                                                          & 2.93                                                                        & 50.49                                                                              \\
\multicolumn{1}{c|}{}                                                                 & \multicolumn{1}{l|}{\cellcolor[HTML]{DFDFDF}Portuguese} & \cellcolor[HTML]{DFDFDF}4,112 & \cellcolor[HTML]{DFDFDF}4,769                                                & \cellcolor[HTML]{DFDFDF}39.00                                                & \cellcolor[HTML]{DFDFDF}8,778                                                & \cellcolor[HTML]{DFDFDF}14.19                                              & \cellcolor[HTML]{DFDFDF}22.74 & \cellcolor[HTML]{DFDFDF}71.1  & \cellcolor[HTML]{DFDFDF}90.70 & \cellcolor[HTML]{DFDFDF}96.58 & \cellcolor[HTML]{DFDFDF}99.70   & \cellcolor[HTML]{DFDFDF}4.40                                                  & \cellcolor[HTML]{DFDFDF}4.14                                                & \cellcolor[HTML]{DFDFDF}31.35                                                      \\
\multicolumn{1}{c|}{}                                                                 & \multicolumn{1}{l|}{Spanish}        & 28,406                     & 3,745                                                                        & 34.94                                                                        & 6,084                                                                        & 17.82                                                                       & 22.34                         & 66.03                         & 86.71                         & 93.87                         & 99.52                           & 5.10                                                                          & 4.00                                                                        & 33.04                                                                              \\
\multicolumn{1}{c|}{}                                                                 & \multicolumn{1}{l|}{\cellcolor[HTML]{DFDFDF}Russian}  & \cellcolor[HTML]{DFDFDF} 28,272   & \cellcolor[HTML]{DFDFDF}709                                                 & \cellcolor[HTML]{DFDFDF}30.79                                                & \cellcolor[HTML]{DFDFDF}1,516                                                & \cellcolor[HTML]{DFDFDF}9.48                                               & \cellcolor[HTML]{DFDFDF}42.32 & \cellcolor[HTML]{DFDFDF}85.46 & \cellcolor[HTML]{DFDFDF}96.33 & \cellcolor[HTML]{DFDFDF}98.83 & \cellcolor[HTML]{DFDFDF}98.66   & \cellcolor[HTML]{DFDFDF}3.38                                                  & \cellcolor[HTML]{DFDFDF}3.54                                                & \cellcolor[HTML]{DFDFDF}19.24                                                      \\
\multicolumn{1}{c|}{}                                                                 & \multicolumn{1}{l|}{Ukrainian}          & 16,997                 & 611                                                                         & 34.21                                                                        & 1,440                                                                        & 8.56                                                                        & 41.31                         & 86.82                         & 96.64                         & 98.93                         & 98.60                           & 3.35                                                                          & 3.28                                                                        & 20.84                                                                              \\
\multicolumn{1}{c|}{}                                                                 & \multicolumn{1}{l|}{\cellcolor[HTML]{DFDFDF}Panjabi}  & \cellcolor[HTML]{DFDFDF}8,195    & \cellcolor[HTML]{DFDFDF}798                                                 & \cellcolor[HTML]{DFDFDF}41.09                                                & \cellcolor[HTML]{DFDFDF}2,140                                                & \cellcolor[HTML]{DFDFDF}13.48                                              & \cellcolor[HTML]{DFDFDF}31.14 & \cellcolor[HTML]{DFDFDF}77.72 & \cellcolor[HTML]{DFDFDF}32.12 & \cellcolor[HTML]{DFDFDF}96.75 & \cellcolor[HTML]{DFDFDF}98.31   & \cellcolor[HTML]{DFDFDF}3.46                                                  & \cellcolor[HTML]{DFDFDF}5.24                                                & \cellcolor[HTML]{DFDFDF}47.41                                                      \\
\multicolumn{1}{c|}{}                                                                 & \multicolumn{1}{l|}{Gujarati}           & 7,218                 & 832                                                                         & 51.83                                                                        & 2,284                                                                        & 10.80                                                                      & 41.61                         & 83.25                         & 94.99                         & 98.06                         & 98.70                           & 3.35                                                                          & 5.33                                                                        & 42.74                                                                              \\
\multicolumn{1}{c|}{}                                                                 & \multicolumn{1}{l|}{\cellcolor[HTML]{DFDFDF}Hindi}   & \cellcolor[HTML]{DFDFDF} 7,191     & \cellcolor[HTML]{DFDFDF}1,251                                                & \cellcolor[HTML]{DFDFDF}65.17                                                & \cellcolor[HTML]{DFDFDF}2,579                                                & \cellcolor[HTML]{DFDFDF}13.10                                              & \cellcolor[HTML]{DFDFDF}23.48 & \cellcolor[HTML]{DFDFDF}68.09 & \cellcolor[HTML]{DFDFDF}88.58 & \cellcolor[HTML]{DFDFDF}95.82 & \cellcolor[HTML]{DFDFDF}98.95   & \cellcolor[HTML]{DFDFDF}3.36                                                  & \cellcolor[HTML]{DFDFDF}4.35                                                & \cellcolor[HTML]{DFDFDF}65.31                                                      \\
\multicolumn{1}{c|}{}                                                                 & \multicolumn{1}{l|}{Marathi}            & 9,396                 & 803                                                                         & 56.55                                                                        & 2,137                                                                        & 9.82                                                                       & 44.75                         & 83.87                         & 95.81                         & 98.84                         & 98.77                           & 3.39                                                                          & 5.28                                                                        & 43.95                                                                              \\
\multicolumn{1}{c|}{}                                                                 & \multicolumn{1}{l|}{\cellcolor[HTML]{DFDFDF}Bengali} & \cellcolor[HTML]{DFDFDF}12,954     & \cellcolor[HTML]{DFDFDF}530                                                 & \cellcolor[HTML]{DFDFDF}36.51                                                & \cellcolor[HTML]{DFDFDF}1,388                                                & \cellcolor[HTML]{DFDFDF}9.49                                               & \cellcolor[HTML]{DFDFDF}35.17 & \cellcolor[HTML]{DFDFDF}81.49 & \cellcolor[HTML]{DFDFDF}94.66 & \cellcolor[HTML]{DFDFDF}98.20 & \cellcolor[HTML]{DFDFDF}98.20   & \cellcolor[HTML]{DFDFDF}3.41                                                  & \cellcolor[HTML]{DFDFDF}3.43                                                & \cellcolor[HTML]{DFDFDF}60.32                                                      \\
\multicolumn{1}{c|}{\multirow{-11}{*}{{\begin{tabular}[c]{@{}c@{}}Indo-\\ European\end{tabular}}}}                                 & \multicolumn{1}{l|}{French}           & 6,344                   & 575                                                                         & 20.23                                                                        & 1,115                                                                        & 9.99                                                                       & 29.72                         & 74.50                         & 91.23                         & 96.32                         & 98.26                           & 3.15                                                                          & 3.42                                                                        & 28.98                                                                              \\
\multicolumn{1}{c|}{Turkic}                                                           & \multicolumn{1}{l|}{\cellcolor[HTML]{DFDFDF}Turkish}  & \cellcolor[HTML]{DFDFDF}5,031   & \cellcolor[HTML]{DFDFDF}556                                                 & \cellcolor[HTML]{DFDFDF}26.72                                                & \cellcolor[HTML]{DFDFDF}1,225                                                & \cellcolor[HTML]{DFDFDF}10.29                                              & \cellcolor[HTML]{DFDFDF}39.70 & \cellcolor[HTML]{DFDFDF}80.53 & \cellcolor[HTML]{DFDFDF}93.55 & \cellcolor[HTML]{DFDFDF}97.53 & \cellcolor[HTML]{DFDFDF}98.14   & \cellcolor[HTML]{DFDFDF}2.28                                                  & \cellcolor[HTML]{DFDFDF}3.46                                                & \cellcolor[HTML]{DFDFDF}44.67                                                      \\
\multicolumn{1}{c|}{Afro-Asiatic}                                                     & \multicolumn{1}{l|}{Arabic}                  & 6,922            & 653                                                                         & 65.80                                                                        & 1,499                                                                        & 11.43                                                                      & 36.35                         & 81.04                         & 94.16                         & 97.88                         & 98.25                           & 2.60                                                                          & 4.47                                                                        & 46.65                                                                              \\
\multicolumn{1}{c|}{Sino-Tibetan}                                                     & \multicolumn{1}{l|}{\cellcolor[HTML]{DFDFDF}Mandarin}  & \cellcolor[HTML]{DFDFDF}12,279  & \cellcolor[HTML]{DFDFDF}1,266                                                & \cellcolor[HTML]{DFDFDF}42.66                                                & \cellcolor[HTML]{DFDFDF}1,429                                                & \cellcolor[HTML]{DFDFDF}13.87                                              & \cellcolor[HTML]{DFDFDF}21.13 & \cellcolor[HTML]{DFDFDF}70.48 & \cellcolor[HTML]{DFDFDF}88.66 & \cellcolor[HTML]{DFDFDF}95.37 & \cellcolor[HTML]{DFDFDF}98.90   & \cellcolor[HTML]{DFDFDF}4.77                                                  & \cellcolor[HTML]{DFDFDF}4.69                                                & \cellcolor[HTML]{DFDFDF}44.94                                                      \\
\multicolumn{1}{c|}{}                                                                 & \multicolumn{1}{l|}{Telugu}            & 9,579                  & 536                                                                         & 51.50                                                                        & 1,568                                                                        & 9.02                                                                       & 50.95                         & 87.51                         & 96.80                         & 98.99                         & 98.31                           & 2.69                                                                          & 5.07                                                                        & 38.42                                                                              \\
\multicolumn{1}{c|}{\multirow{-2}{*}{Dravidian}}                                      & \multicolumn{1}{l|}{\cellcolor[HTML]{DFDFDF}Tamil}    & \cellcolor[HTML]{DFDFDF}9,973   & \cellcolor[HTML]{DFDFDF}440                                                 & \cellcolor[HTML]{DFDFDF}34.75                                                & \cellcolor[HTML]{DFDFDF}1,110                                                & \cellcolor[HTML]{DFDFDF}8.84                                               & \cellcolor[HTML]{DFDFDF}47.94 & \cellcolor[HTML]{DFDFDF}86.67 & \cellcolor[HTML]{DFDFDF}96.65 & \cellcolor[HTML]{DFDFDF}99.06 & \cellcolor[HTML]{DFDFDF}97.99   & \cellcolor[HTML]{DFDFDF}2.29                                                  & \cellcolor[HTML]{DFDFDF}4.14                                                & \cellcolor[HTML]{DFDFDF}39.56                                                      \\
\multicolumn{1}{c|}{Austroasiatic}                                                    & \multicolumn{1}{l|}{Nepali}               & 6,185               & 440                                                                         & 18.77                                                                        & 1,178                                                                        & 9.47                                                                       & 44.17                         & 86.63                         & 96.47                         & 99.12                         & 97.85                           & 2.30                                                                          & 3.40                                                                        & 54.70                                                                              \\
\multicolumn{1}{c|}{}                                                                 & \multicolumn{1}{l|}{\cellcolor[HTML]{DFDFDF}Persian}  & \cellcolor[HTML]{DFDFDF}8,830   & \cellcolor[HTML]{DFDFDF}716                                                 & \cellcolor[HTML]{DFDFDF}27.19                                                & \cellcolor[HTML]{DFDFDF}1,352                                                & \cellcolor[HTML]{DFDFDF}11.42                                               & \cellcolor[HTML]{DFDFDF}24.82 & \cellcolor[HTML]{DFDFDF}70.32 & \cellcolor[HTML]{DFDFDF}89.24 & \cellcolor[HTML]{DFDFDF}95.67 & \cellcolor[HTML]{DFDFDF}98.41   & \cellcolor[HTML]{DFDFDF}2.39                                                  & \cellcolor[HTML]{DFDFDF}3.08                                                & \cellcolor[HTML]{DFDFDF}52.37                                                      \\
\multicolumn{1}{c|}{\multirow{-2}{*}{Indo-Iranian}}                                   & \multicolumn{1}{l|}{Urdu}           & 13,469                     & 964                                                                         & 36.84                                                                        & 1,712                                                                        & 14.02                                                                      & 22.55                         & 63.25                         & 82.91                         & 91.29                         & 98.55                           & 2.94                                                                          & 3.88                                                                        & 57.48                                                                              \\
\multicolumn{1}{c|}{Austronesian}                                                     & \multicolumn{1}{l|}{\cellcolor[HTML]{DFDFDF}Indonesian} & \cellcolor[HTML]{DFDFDF}12,951 & \cellcolor[HTML]{DFDFDF}779                                                 & \cellcolor[HTML]{DFDFDF}36.74                                                & \cellcolor[HTML]{DFDFDF}1,432                                                & \cellcolor[HTML]{DFDFDF}11.38                                              & \cellcolor[HTML]{DFDFDF}29.34 & \cellcolor[HTML]{DFDFDF}76.65 & \cellcolor[HTML]{DFDFDF}93.18 & \cellcolor[HTML]{DFDFDF}97.78 & \cellcolor[HTML]{DFDFDF}98.53   & \cellcolor[HTML]{DFDFDF}4.82                                                  & \cellcolor[HTML]{DFDFDF}3.04                                                & \cellcolor[HTML]{DFDFDF}34.55                                                      \\ \Xhline{4\arrayrulewidth}
\multicolumn{2}{c}{\textbf{Summary}}          & \textbf{415,117}                                                                                                   & \textbf{902}                                                                & \textbf{27.17}                                                               & \textbf{1,632}                                                               & \textbf{10.13}                                                              & \textbf{33.60}                & \textbf{80.83}                & \textbf{94.37}                & \textbf{98.89}                & \textbf{98.88}                  & \textbf{3.21}                                                                 & \textbf{3.47}                                                               & \textbf{44.64}    \\ \Xhline{4\arrayrulewidth}                                                             
\end{tabular}}
\caption{Statistics about our \texttt{\textbf{XL-HeadTags}}. `\#Samples' denotes the total number of samples. `Avg. Word,' `Avg. Sent,' and `Avg. Tok' represent the average number of words, sentences, and tokens per document (computed using the BERT-multilingual model \cite{devlin2019bert}), respectively. `\% of Novel N-grams' (for n = 1, 2, 3, 4) indicates the proportion of novel n-grams in headlines. `CR' refers to the compression ratio of headlines. 
`Avg. I/C' shows the average number of image-caption pairs per sample. `Avg. Tag W' signifies the average number of tag words per document. `\% Pre Tag W' denotes the average percentage of present tag words in the documents.}
\label{table:datas-stats-1}
\end{table*}

\subsection{Tasks}
\label{sec:XL-HeadTags}

Our goal is to simultaneously generate headline and tags for news articles. Given that headlines provide a condensed summary and tag words serve as semantic markers, we propose that these two tasks can be effectively learned in a unified learning framework. Formally, Given a news Article $\mathcal{A}$ consisting of sequence of words $\{a_1, a_2, \cdots , a_n\}$, our objective is to generate an abstractive Headline, $\mathcal{H}$ consisting of sequence of words $\{h_1, h_2, \cdots , h_m\}$ and a set of $\mathcal{T}$ Tag Words $\{ \{T_1\}, \{T_2\}, \cdots , \{T_o\} \}$, where each $T$ can be a word $T = \{t_1\}$ or a sequence of word $T = \{t_1, t_2, \cdots , t_p\}$. Thus the task is $\textbf{XL-HeadTags}(\mathcal{A}) \implies  \mathcal{H}, \mathcal{T}$.

\paragraph{Controlled Generation}
\label{sec:controlled-generation}
During the tag word generation phase, we introduce an additional setting to control the number of tag word generations. This task is formally defined as $\textbf{Con-XL-HeadTags}(\mathcal{A}, \mathcal{N}) \implies  \mathcal{H}, \mathcal{T}_{1:n}^{con}$. Here, $\mathcal{N}$ acts as the control factor determining the number of tag words to be generated, and $\mathcal{T}_{1:n}^{con}$ represents the resulting set of $n$ controlled tag words.

\section{MultiRAGen}
\label{sec:task-definition-MultiRAGen}

Retrieval-Augmented Generation (\textbf{RAG}) \cite{10.5555/3495724.3496517} involves two key phases: \textbf{first}, retrieving contextually relevant information, and \textbf{second}, using this information to guide the generation process \cite{zhao-etal-2023-retrieving}. RAG has been applied to various tasks such as \emph{machine translation} \cite{he-etal-2021-fast}, \emph{dialogue generation} \cite{cai-etal-2019-skeleton}, and \emph{abstractive summarization} \cite{peng-etal-2019-text}. Inspired by these applications, we leverage multimodal information like images and captions to select salient content from news articles, introducing \textbf{MultiRAGen} (\textbf{Multi}modal \textbf{R}etrieval \textbf{A}ugmented \textbf{Gen}eration). MultiRAGen comprises two main components: \textbf{(1)} Multimodal Retrievers (\secref{sec:retriever}) and \textbf{(2)} Instruction Tuning (\secref{sec:instruction-tuning}).

\subsection{Multimodal Retrievers (MultiRet)}
\label{sec:retriever}

Our approach utilizes auxiliary information, such as images and captions, to extract salient and relevant sentences from documents through two modules: \texttt{\textbf{ImgRet}} for images (\texttt{visual modality}) and \texttt{\textbf{CapRet}} for image captions (\texttt{textual modality}). Formally, these modules are described as follows:
\begin{equation*}
\resizebox{.98\columnwidth}{!}{%
$\texttt{\textbf{MultiRet}} = 
\left\{
    \begin{array}{l}
        \texttt{\textbf{ImgRet}}(\mathcal{A}_{1:n'}, \mathcal{I}_{1:m'}, \mathcal{K}, \mathcal{L}) \implies \mathcal{A}_{1:k}^I, \\ \\
        \texttt{\textbf{CapRet}}(\mathcal{A}_{1:n'}, \mathcal{C}_{1:m'}, \mathcal{K}, \mathcal{L}) \implies \mathcal{A}_{1:k}^C
    \end{array}
\right.$%
}
\end{equation*}
where $\mathcal{A}_{1:n'}$ represents the article consisting of $n'$ sentences, $\mathcal{I}_{1:m'}$ and $\mathcal{C}_{1:m'}$ denote the set of images and image captions within the document, respectively, with $m' \geqslant 1$. $\mathcal{K}$ is the number of sentences to be retrieved, $\mathcal{A}_{1:k}^I$ and $\mathcal{A}_{1:k}^C$ are the resulting subsets containing $k$ sentences selected based on visual and textual modalities, respectively. Here, $k=\mathcal{K}$ if $\mathcal{K} \leqslant n'$, otherwise $k=n'$. $\mathcal{L}$ specifies the language of the article.

We tokenize the documents into sentences using our \texttt{Multilingual Sentence Tokenizer} (introduced later in Section \ref{sec:tools}) and use images and captions as queries to compute semantic similarity with the sentences. This process employs a multilingual version of the \texttt{\textbf{OpenAI CLIP-ViT-B32}} model\footnote{\href{https://huggingface.co/sentence-transformers/clip-ViT-B-32-multilingual-v1}{clip-ViT-B-32-multilingual-v1}} \cite{pmlr-v139-radford21a}, which maps text (\emph{in 50+ languages}) and images to a shared dense vector space \cite{reimers-2019-sentence-bert}, aligning images closely with their corresponding texts. We then retrieve the top $\mathcal{K}$ sentences from the document based on the similarity scores.

\paragraph{Handling Multiple Images and Captions} We observed that a single document often contains multiple images and captions without a proper one-to-one mapping between them. Consequently, we treat each image and caption as distinct entities and propose a greedy algorithm for aggregating multiple retrievals. The detailed procedural depiction of one this simple yet effective algorithmic processes is presented in Algorithm~\ref{algo:ImgRet} and detailed in Appendix ~\ref{sec:greddy-algo}.

\begin{algorithm}[t]
  \caption{ImgRet
  \label{algo:ImgRet}}
  \Function{ImgRet($Article$, $L$, $Images$, $K$)}{
    $\var{sentences} \assign \FuncCall{SenSeg}{$Article$, $L$} $
    \For{sen \textbf{in} sentences}{             
        \For{img \textbf{in} Images}{
            $\var{scr} \assign \FuncCall{SimScr}{$sen$, $img$}$
            
            $\var{sim} \assign \var{sim} + \var{scr}$
        }
    }
    \Comment{Sort the sentences on \var{sim}}
    
    $\var{article} \assign \var{sentences[1:K]} $

    \Comment{Sort the sentences on order}
    
    \Return{article}
  }  
\end{algorithm}

\subsection{Instruction Tuning}
\label{sec:instruction-tuning}

Language models can be fine-tuned with supervised datasets containing natural language prompts and their corresponding target completions \cite{wei2022finetuned, sanh2022multitask, NEURIPS2022-b1efde53, min-etal-2022-metaicl}. This process, known as \emph{``instruction tuning,''} enhances the models' ability to follow instructions accurately. Typically, task-specific prefixes are used to guide the model towards the desired output format. Inspired by the adaptability and success of these approaches in managing diverse tasks via a unified text-to-text framework, we apply this methodology to generate headlines and tags for news articles within a multilingual setting. We introduce two instructional variations: one for \textbf{unrestricted} and another for \textbf{controlled} tag word generation along with headline.

\paragraph{Unrestricted Generation}
\label{sec:unrestricted-generation-instruction}
allows the model to independently determine the optimal number of tag words to generate. In the following input instruction, \textbf{bold} indicates the input prefix, while \underline{$underline$} signifies the selected content (described below). Conversely, in the output instruction, \textbf{bold} marks the output prefix, and \underline{$underline$} denotes the generated output, encompassing both the headline and a variable number of tag words ($T_1$, $T_2$, ...) corresponding to each article. This approach aims to simultaneously address two text generation tasks, utilizing output prefixes to distinguish between the outcomes of each task. The complete instruction format is as follows:

\begin{tcolorbox}[breakable,title={\small Instruction for Unrestricted Generation}]
\footnotesize
\textcolor{darkgreen}{\texttt{Input}} \ding{221} \textbf{Generate Headline and Tag Words: \underline{$Selected \ Content$}.} \\
\vspace{1mm}

\textcolor{darkgreen}{\texttt{Output}} \ding{221} \textbf{Headline is: \underline{$Headline$}. Tag words are: \underline{$T_1$, $T_2$, $\cdots$, $T_o$}}.
\end{tcolorbox}

\paragraph{Controlled Generation}
\label{sec:controlled-generation-instruction}
To tackle the challenge of determining the correct number of tag words, we have adjusted our unrestricted prefix. This modification allows us to directly specify the desired number of tag words in the prefix. Our revised input format for controlled tag word generation is: 

\begin{tcolorbox}[breakable,title={\small Instruction for Controlled Generation}]
\footnotesize
\textcolor{darkgreen}{\texttt{Input}} \ding{221} \textbf{Generate Headline and $\mathcal{N}$ Tag Words: \underline{$Selected \ Content$}.}
\end{tcolorbox}

Here, $\mathcal{N}$ refers to the number of tag words to generate. During model training, this number is represented as the count of tag words associated with the original article from the training dataset, verbalized in natural language (such as One, Two, Three, ...). The output format remains unchanged.

\paragraph{\underline{Selected Content}} For the input format of the instruction, we utilize three settings: \textbf{(1)} Article, where the original article is placed; \textbf{(2)} Top $\mathcal{K}$ sentences retrieved by our \texttt{\textbf{MultiRet}}; and \textbf{(3)} Top $\mathcal{K}$ sentences from our \texttt{\textbf{MultiRet}} concatenated with the Article. \texttt{\textbf{MultiRet}} comprises two modules: \texttt{\textbf{ImgRet}}, which retrieves relevant sentences using images as queries, and \texttt{\textbf{CapRet}}, which uses image captions as queries to retrieve pertinent sentences from the input article as detailed in Section \secref{sec:retriever}.

\section{Multilingual Tools}
\label{sec:tools}

We also develop tools by accumulating open-source resources for processing and evaluating multilingual texts, making it an easy-to-use one-stop destination. These tools include \textbf{(1)} Multilingual ROUGE Scorer, \textbf{(2)} Multilingual Sentence Tokenizer, and \textbf{(3)} Multilingual Stemmer. These resources are invaluable to the research community focusing on multilingual NLP, providing essential support for accurate processing and evaluation\footnote{These tools are available here \href{https://github.com/faisaltareque/XL-HeadTags}{XL-HeadTags}, and can be easily installed via \texttt{`pip'}.} (also detailed in Appendix \ref{sec:multilingual-tools}).

\paragraph{\texttt{Multilingual ROUGE Scorer}} \citet{hasan-etal-2021-xl, chronopoulou2023language} identified a significant issue in evaluating multilingual summarization performance: the absence of stemmers for certain low-resource languages hindered the processing of generated summaries, resulting in lower ROUGE scores \cite{lin-2004-rouge}. Facing a similar challenge, \citet{aharoni-etal-2023-multilingual} calculated ROUGE scores using a multilingual tokenizer. However, the lack of word tokenizers for some languages still presents a challenge for fair ROUGE score assessment across different language families. To address this issue, we developed a Multilingual ROUGE Scorer that leverages Byte-Pair Encoding (BPE) tokenization from BERT-multilingual \cite{devlin2019bert}, ensuring more accurate evaluation across \num{104} languages.

\paragraph{\texttt{Multilingual Sentence Tokenizer}} Sentence tokenization aims to divide a given document into individual sentences. To achieve this, we integrated various open-source resources for multiple languages into a unified library. This tool can perform tokenization in \num{41} different languages (details of the sources in Appendix (Table \ref{table:multilingual-sentence-tokenizer})).

\paragraph{\texttt{Multilingual Stemmer}} For evaluating the performance of keyphrase generation, both generated and reference keyphrases are normalized before assessing exact matches \cite{meng-etal-2017-deep, chen-etal-2020-exclusive}. Given the need to evaluate generated tag words across various languages, we developed a Multilingual Stemmer that integrates open-source stemmers for \num{18} distinct languages (as detailed in Table \ref{table:multilingual-stemmer} in the Appendix). However, open-source stemmers for \textbf{Chinese} and \textbf{Telugu} are unavailable. Therefore, when evaluating tag words in these languages, we report the scores without normalizing the tag words.

\section{Evaluating Tags Generation}
\label{sec:tag-gen-metric}

\begin{table*}[t]
\centering
\renewcommand{\arraystretch}{1} 
\resizebox{14cm}{!} 
{ 
\begin{tabular}{ccccccccc}
\Xhline{4\arrayrulewidth}
\multicolumn{2}{c}{}                                                                                     & \textbf{R1}                                          & \textbf{R2}                                          & \textbf{RL}                                          & \textbf{BLEU}                                        & \textbf{METEOR}                                      & \textbf{LR} ($\downarrow$)                                         & \textbf{BERT Score}                                       \\ \hline \hline
\multicolumn{2}{c|}{\textbf{Models}}                                                                     & \multicolumn{7}{c}{\textbf{Baselines}}                                                                                                                                                                                                                                                                                                                                                        \\ \hline 
\rowcolor[HTML]{DCDCDC} 
\multicolumn{2}{c|}{\cellcolor[HTML]{DCDCDC}mT5}                                                         & 37.86                                                & 17.20                                                & 33.53                                                & 12.95                                                & 25.55                                                & 0.84                                                & 75.79                                                \\
\end{tabular}
}
\caption{Headline Generation Evaluation. \textbf{Selected Content} (Important Sentences + Article). \textbf{K} denotes the number of sentences retrieved for both text and visual modalities. ($\downarrow$) indicates lower values for better performance. The best results compared to their respective baseline models are marked in \textbf{bold}, and $\Delta$ gains are shown in round brackets and highlighted with \textcolor{darkgreen}{green} and \textcolor{customRed}{red} colors.}
\label{table:headline-evaluation-ia}
\end{table*}

The terms \emph{``Tag Words''} and \emph{``Keyphrases,''} while sharing a common objective, differ in their application. In the keyphrase generation process, a model predicts a set of distinct keyphrases $\mathcal{\hat Y} = (\hat y_1, \dots, \hat y_m)$ from a given source text, with these predictions $\hat y_i$ being ordered based on their relevance \cite{yuan-etal-2020-one}. The ground truth keyphrases for the source text are denoted as $\mathcal{Y}$. It's important to note that in the context of tag words, the order of predictions does not necessarily reflect the quality of each prediction. To measure predictive performance, three standard evaluation metrics—namely macro-averaged precision, recall, and F-measure ($F_1$)—are commonly used \cite{meng-etal-2017-deep}. 
Formally, the metrics of precision, recall, and $F_1$ score are defined as follows:

\begin{equation}
\begin{aligned}
    P &= \frac{|\mathcal{\hat Y}\cap\mathcal{Y}|}{|\mathcal{\hat Y}|}, 
    \:\:\:\:\:\:\:\:
    R = \frac{|\mathcal{\hat Y}\cap\mathcal{Y}|}{|\mathcal{Y}|},   \\
    F_1 &= \frac{2 * P * R}{P + R}.
\end{aligned}
\end{equation}

\subsection{Proposed Tag Words Evaluation Metrics}
\label{sec:proposed-tag-eval}
As detailed in Section \ref{sec:unrestricted-generation-instruction}, our work spans both controlled and unrestricted tag word generation. Inspired by \citet{yuan-etal-2020-one}, we introduce three metrics for assessing performance. 

\paragraph{Unrestricted Generation}
In unrestricted generation, a varying number of tag words are generated. The evaluation employs the metric $F_{1}@\mathcal{M}$, where $|\mathcal{\hat Y}| = \mathcal{M}$. Here, $\mathcal{M}$ varies with each article, reflecting the model's autonomous decision on the number of tag words.

\paragraph{Controlled Generation}
For controlled generation, the goal is to generate a fixed number of tag words. We evaluate this using $F_{1}@\mathcal{K}$ and $F_{1}@\mathcal{O}$. $\mathcal{K}$ is predefined as 3 and 5, and $\mathcal{O}$ corresponds to $|\mathcal{Y}|$, the actual number of tag words. This means we assess controlled generation for producing exactly 3, 5, and $|\mathcal{Y}|$ tag words, as specified by a natural language prefix.

\section{Experiments}

\subsection{Data \& Evaluation Metrics}

\paragraph{Data} To ensure a balanced distribution, our dataset division for training, validation, and test sets across all languages consists of 95\% (394,353 samples), 1\% (5,187 samples), and 4\% (15,577 samples) respectively from our \texttt{\textbf{XL-HeadTags}} dataset (detailed in Appendix (Table \ref{table:datas-stats-1})). Our goal is to enhance the task's generalizability by developing a unified model capable of performing both \textbf{controlled} (\emph{a fixed number of tag words}) and \textbf{unrestricted} (\emph{the model decides the number of tag words}) tag word generation (also elaborated in Section \ref{sec:instruction-tuning}). To achieve this, we introduce a \textbf{prefix mixture strategy} during training, using a 70:30 allocation ratio. Here, 70\% of the data is formatted for controlled tag word generation, while the remaining 30\% is free from such constraints. This approach, applied consistently across training, validation, and test datasets for both baseline and our models, aims to facilitate a model adept at navigating both task variations. Details on data processing are presented in Appendix \ref{sec:dataset-processing}.

\paragraph{Evaluation Metrics} For the evaluation of generated headlines, we utilize F1 score of our \texttt{Multilingual ROUGE Scorer} (\secref{sec:tools}) and the BLEU score \citep{Papineni02bleu:a}, F1 BERT score \citep{bert-score}, METEOR score \citep{banarjee2005}, and Length Ratio (LR). For assessing tag words, we apply the metrics we have proposed in Section \ref{sec:proposed-tag-eval} and normalize the tag words using our \texttt{Multilingual Stemmer} (\secref{sec:tools}).

\subsection{Models}

\paragraph{Baselines} We use \texttt{mT5-base} \cite{xue-etal-2021-mt5}, \texttt{mT0-base} \cite{mt0-muennighoff2022crosslingual} and \texttt{Flan-T5-large} \cite{chung2022scaling} checkpoints available in the Hugging Face \cite{wolf-etal-2020-transformers}.
Notably, these models are multilingual and pretrained on the mC4\footnote{\url{https://huggingface.co/datasets/mc4}} multilingual corpus. It is essential to highlight that the mT0 model underwent additional fine-tuning within a multitask framework, utilizing the crosslingual task mixture xP3 \cite{mt0-muennighoff2022crosslingual} dataset to enhance crosslingual generalization. We conducted fine-tuning of the mT5-base, mT0-base and Flan-T5-large models using the original article in the instruction. Additional baselines like LEAD-1 and EXT-ORACLE are detailed in Appendix \ref{sec:baselines}.  


\paragraph{Gemini Pro and Mixtral} We employed the Gemini Pro \cite{geminiteam2023gemini} and Mixtral \cite{jiang2023mistral} models for evaluating their efficacy in \textbf{XL-HeadTags} multilingual tasks. This assessment occurred in zero-shot prompting conditions, with instructions specifying input (\textit{i.e., article}) and output formats (\secref{sec:instruction-tuning}). This encompassed sampling \num{50} instances from each language.

\paragraph{MultiRAGen (\textit{ours})} For visual modality, we use images, and for textual modality, we utilize image captions. The number of sentences to retrieve is determined by the parameter $\mathcal{K}$, with explored values of 5, 10, and 15, corresponding to the retrieval of the top 5, 10, and 15 sentences, respectively. After retrieval, we reorder the top-$\mathcal{K}$ sentences to their original sequence in the article, ensuring the narrative flow remains coherent. Our experimentation includes two approaches: \textbf{(1)} inputting only the top-$\mathcal{K}$ retrieved sentences and \textbf{(2)} combining these sentences with the original article. We apply the same set of hyperparameters for all the baselines and our models, as detailed in Appendix \ref{sec:hyperparameter}.

\section{Results and Discussion}
\label{sec:results-discussions}

The evaluation results for headline generation are detailed in Table \ref{table:headline-evaluation-ia}, and the outcomes for tag word evaluation are shown in Table \ref{table:tags-evaluation}. We also compare the performance of our baseline models with three extractive methods—\textbf{TF-IDF} \cite{tfidf} and \textbf{TextRank} \cite{mihalcea-tarau-2004-textrank} as unsupervised, and \textbf{KEA} \cite{witten1999kea} as supervised—on the English test set. These extractive methods are implemented using the \textbf{PKE} module \cite{boudin:2016:COLINGDEMO}, with a detailed comparison presented in Table \ref{table:tags-evaluation=other}.

\subsection{Headline}
\paragraph{Baselines} Table \ref{table:headline-evaluation-ia} shows LEAD-1's poor performance, indicating its inability to capture the abstractive essence of headlines. EXT-ORACLE, however, provides a robust baseline with improved ROUGE and BLEU scores, though it lacks headline conciseness. mT5, mT0, and Flan-T5 outperform extractive methods, with mT0 closely matching the conciseness of reference headlines. Flan-T5 excels in high-resource languages but  falls short in low-resource setting (also in Table \ref{table:language-headline-flant5}, and \ref{table:language-headline-flant5-ia}).

\paragraph{MultiRAGen} results show that using captions for retrieval with mT5 outperforms the standard mT5 baseline, a pattern also seen with mT0 and Flan-T5. This indicates that including the most relevant information in the context window enhances headline generation. Similarly, using images for retrieval with mT5, mT0, and Flan-T5 also surpasses their respective baselines, performing nearly as well as caption-based retrieval. This suggests potential benefits from integrating both textual and visual modalities in future research to improve retrieval effectiveness (See Limitations (\secref{sec:limitations})).

\begin{table}[t]
\centering
\renewcommand{\arraystretch}{1} 
\resizebox{5.95cm}{!} 
{ 
\begin{tabular}{ccccc}
\Xhline{4\arrayrulewidth}
\textbf{Models}            & \ft & \ff & \fm & \fo \\ \hline
\multicolumn{5}{c}{\textbf{Extractive Unsupervised}} \\ \hline
\rowcolor[HTML]{E8E8E8} 
TF-IDF & 10.9 & 11.4 & - & 10.85 \\
TextRank & 0.71 & 1.05 & - & 0.74 \\ \Xhline{4\arrayrulewidth}
\multicolumn{5}{c}{\textbf{Extractive Supervised}} \\ \hline
\rowcolor[HTML]{E8E8E8} 
KEA & 10.55 & 11.04 & - & 10.87 \\ \Xhline{4\arrayrulewidth}
\multicolumn{5}{c}{\textbf{Abstractive}} \\ \hline
\rowcolor[HTML]{E8E8E8} 
Gemini-Pro & 18.01 & 21.12 & 19.22 & 18.37 \\
Mixtral & 7.20 & 10.69 & 7.53 & 8.54 \\ \Xhline{4\arrayrulewidth}
\multicolumn{5}{c}{\textbf{Abstractive Baselines-Ours}} \\ \hline
\rowcolor[HTML]{E8E8E8} 
mT5 & 44.51 & 37.91 & 45.34 & 47.80 \\
mT0 & 51.52 & 43.57 & 54.36 & 57.00 \\
\rowcolor[HTML]{E8E8E8} 
Flan-T5 & 50.2 & 43.17 & 53.0 & 55.37 \\
\Xhline{4\arrayrulewidth}
\end{tabular}
}
\caption{Tag Words Evaluation on English. }
\label{table:tags-evaluation=other}
\end{table}

\begin{table*}[t]
\centering
\renewcommand{\arraystretch}{1} 
\resizebox{10.65cm}{!} 
{ 
\begin{tabular}{cccccc}
\Xhline{4\arrayrulewidth}
\multicolumn{2}{c}{}                                                                                    & \ft            & \ff            & \fm            & \fo            \\ \Xhline{4\arrayrulewidth}
\multicolumn{2}{c}{\textbf{Models}}                                                                     & \multicolumn{4}{c}{\textbf{Baselines}}                                                                                        \\ \hline
\rowcolor[HTML]{E8E8E8} 
\multicolumn{2}{c|}{\cellcolor[HTML]{E8E8E8}mT5}                                                        & 45.01                         & 39.82                         & 44.67                         & 46.79                         \\
\multicolumn{2}{c|}{mT0}                                                                                & 51.58                         & 44.94                         & 52.50                         & 54.39                         \\
\rowcolor[HTML]{E8E8E8} 
\multicolumn{2}{c|}{\cellcolor[HTML]{E8E8E8}Flan-T5}                                                                                & 30.76                         & 26.3                         & 31.86                         & 33.4                         \\ 

\multicolumn{2}{c|}{Gemini-Pro}                                                                          & 5.90                                                    & 7.75                                                    & 6.57                                                    & 6.65                                                    \\
\rowcolor[HTML]{DCDCDC} 
\multicolumn{2}{c|}{\cellcolor[HTML]{DCDCDC}Mixtral}                                                     & 1.93                                                    & 2.90                                                    & 2.49                                                    & 2.28                                                    \\

\Xhline{4\arrayrulewidth}
\textbf{Modality}                             & \textbf{Models}                                          & \multicolumn{4}{c}{\textbf{MultiRAGen (ours)}}                                                                                \\ \hline
\multicolumn{1}{c|}{}                        & \multicolumn{1}{l|}{\textbf{mT5}}                                 & \multicolumn{4}{c}{}                                                                                                          \\
\multicolumn{1}{c|}{}                        & \multicolumn{1}{c|}{\cellcolor[HTML]{E8E8E8} \enspace \ThickVLine -- w/C (\textbf{K=5})}  & \cellcolor[HTML]{E8E8E8}51.16 (\textcolor{darkgreen}{+\num{6.15}}) & \cellcolor[HTML]{E8E8E8}45.18 (\textcolor{darkgreen}{+\num{5.36}}) & \cellcolor[HTML]{E8E8E8}52.20 (\textcolor{darkgreen}{+\num{7.53}}) & \cellcolor[HTML]{E8E8E8}54.36 (\textcolor{darkgreen}{+\num{7.57}})  \\

\multicolumn{1}{c|}{}                        & \multicolumn{1}{c|}{ \; \ \enspace \ThickVLine -- w/C \textbf{(K=10)}}                          & \textbf{53.08} (\textcolor{darkgreen}{+\num{8.07}}) & \textbf{47.00} (\textcolor{darkgreen}{+\num{7.18}}) & \textbf{54.00} (\textcolor{darkgreen}{+\num{9.33}}) & \textbf{56.24} (\textcolor{darkgreen}{+\num{9.45}}) \\

\multicolumn{1}{c|}{}                        & \multicolumn{1}{c|}{\cellcolor[HTML]{E8E8E8}  \; \enspace \ThickVLine -- w/C (\textbf{K=15})} & \cellcolor[HTML]{E8E8E8}47.66 (\textcolor{darkgreen}{+\num{2.65}}) & \cellcolor[HTML]{E8E8E8}42.56 (\textcolor{darkgreen}{+\num{2.74}}) & \cellcolor[HTML]{E8E8E8}47.37 (\textcolor{darkgreen}{+\num{2.7}}) & \cellcolor[HTML]{E8E8E8}49.96 (\textcolor{darkgreen}{+\num{3.17}}) \\

\multicolumn{1}{c|}{}                        & \multicolumn{1}{l|}{\textbf{mT0}}                                 & \multicolumn{4}{c}{}                                                                                                          \\
\multicolumn{1}{c|}{}                        & \multicolumn{1}{c|}{\cellcolor[HTML]{E8E8E8} \enspace \ThickVLine -- w/C \textbf{(K=5)}}  & \cellcolor[HTML]{E8E8E8}52.10 (\textcolor{darkgreen}{+\num{0.52}}) & \cellcolor[HTML]{E8E8E8}46.41 (\textcolor{darkgreen}{+\num{1.47}}) & \cellcolor[HTML]{E8E8E8}53.50 (\textcolor{darkgreen}{+\num{1.00}}) & \cellcolor[HTML]{E8E8E8}55.62 (\textcolor{darkgreen}{+\num{1.23}}) \\

\multicolumn{1}{c|}{}                        & \multicolumn{1}{c|}{ \; \  \enspace \ThickVLine -- w/C \textbf{(K=10)}}                        & 53.88 (\textcolor{darkgreen}{+\num{2.30}})  & 47.95 (\textcolor{darkgreen}{+\num{3.01}}) & \textbf{55.29} (\textcolor{darkgreen}{+\num{2.79}})                         & \textbf{57.49} (\textcolor{darkgreen}{+\num{3.10}})                         \\

\multicolumn{1}{c|}{}  & \multicolumn{1}{c|}{\cellcolor[HTML]{E8E8E8} \; \enspace \ThickVLine -- w/C \textbf{(K=15)}} & \cellcolor[HTML]{E8E8E8}\textbf{54.05} (\textcolor{darkgreen}{+\num{2.47}}) & \cellcolor[HTML]{E8E8E8}\textbf{48.19} (\textcolor{darkgreen}{+\num{3.25}}) & \cellcolor[HTML]{E8E8E8}55.18 (\textcolor{darkgreen}{+\num{2.68}}) & \cellcolor[HTML]{E8E8E8}57.36 (\textcolor{darkgreen}{+\num{2.97}}) \\

\multicolumn{1}{c|}{}                        & \multicolumn{1}{l|}{\textbf{Flan-T5}}                                 & \multicolumn{4}{c}{}                                                                                                          \\
\multicolumn{1}{c|}{}                        & \multicolumn{1}{c|}{\cellcolor[HTML]{E8E8E8} \enspace \ThickVLine -- w/C \textbf{(K=5)}}  & \cellcolor[HTML]{E8E8E8}30.58 (\textcolor{customRed}{-\num{0.18}}) & \cellcolor[HTML]{E8E8E8}26.12 (\textcolor{customRed}{-\num{0.18}}) & \cellcolor[HTML]{E8E8E8}31.48 (\textcolor{customRed}{-\num{0.38}}) & \cellcolor[HTML]{E8E8E8}32.96 (\textcolor{customRed}{-\num{0.44}}) \\

\multicolumn{1}{c|}{}                        & \multicolumn{1}{c|}{ \; \ \enspace \ThickVLine -- w/C \textbf{(K=10)}}                          & 31.18 (\textcolor{darkgreen}{+\num{0.42}}) & 26.65 (\textcolor{darkgreen}{+\num{0.35}}) & 32.16 (\textcolor{darkgreen}{+\num{0.3}}) & 33.77 (\textcolor{darkgreen}{+\num{0.37}})                        \\

\multicolumn{1}{c|}{\multirow{-12}{*}{{\textbf{Text} (\textit{Caption})}}}  & \multicolumn{1}{c|}{\cellcolor[HTML]{E8E8E8}\; \enspace \ThickVLine -- w/C \textbf{(K=15)}} & \cellcolor[HTML]{E8E8E8}\textbf{31.48} (\textcolor{darkgreen}{+\num{0.72}}) & \cellcolor[HTML]{E8E8E8}\textbf{26.90} (\textcolor{darkgreen}{+\num{0.60}}) & \cellcolor[HTML]{E8E8E8}\textbf{32.4} (\textcolor{darkgreen}{+\num{0.54}}) & \cellcolor[HTML]{E8E8E8}\textbf{34.00} (\textcolor{darkgreen}{+\num{0.60}}) \\ 
\Xhline{4\arrayrulewidth}
\multicolumn{1}{c|}{}                        & \multicolumn{1}{l|}{\textbf{mT5}}                                 & \multicolumn{4}{c}{}  \\

\multicolumn{1}{c|}{}                        & \multicolumn{1}{c|}{\cellcolor[HTML]{E8E8E8} \enspace \ThickVLine -- w/I \textbf{(K=5)}}  & \cellcolor[HTML]{E8E8E8}50.72 (\textcolor{darkgreen}{+\num{5.71}}) & \cellcolor[HTML]{E8E8E8}44.60  (\textcolor{darkgreen}{+\num{4.78}})& \cellcolor[HTML]{E8E8E8}51.70 (\textcolor{darkgreen}{+\num{7.03}}) & \cellcolor[HTML]{E8E8E8}53.52 (\textcolor{darkgreen}{+\num{6.73}}) \\

\multicolumn{1}{c|}{}                        & \multicolumn{1}{c|}{ \; \ \enspace \ThickVLine -- w/I \textbf{(K=10)}}                          & \textbf{53.62} (\textcolor{darkgreen}{+\num{8.61}})  & \textbf{47.57} (\textcolor{darkgreen}{+\num{7.75}}) & \textbf{54.76} (\textcolor{darkgreen}{+\num{10.09}}) & \textbf{56.95} (\textcolor{darkgreen}{+\num{10.16}})                        \\

\multicolumn{1}{c|}{}                        & \multicolumn{1}{c|}{\cellcolor[HTML]{E8E8E8} \; \enspace \ThickVLine -- w/I \textbf{(K=15)}} & \cellcolor[HTML]{E8E8E8}47.67 (\textcolor{darkgreen}{+\num{2.66}}) & \cellcolor[HTML]{E8E8E8}42.39 (\textcolor{darkgreen}{+\num{2.57}}) & \cellcolor[HTML]{E8E8E8}47.21 (\textcolor{darkgreen}{+\num{2.54}}) & \cellcolor[HTML]{E8E8E8}49.80 (\textcolor{darkgreen}{+\num{3.01}}) \\
\multicolumn{1}{c|}{}                        & \multicolumn{1}{l|}{\textbf{mT0}}                                 & \multicolumn{4}{c}{}                                                                                                          \\
\multicolumn{1}{c|}{}                        & \multicolumn{1}{c|}{\cellcolor[HTML]{E8E8E8} \enspace \ThickVLine -- w/I \textbf{(K=5)}}  & \cellcolor[HTML]{E8E8E8}50.85 (\textcolor{customRed}{-\num{0.73}}) & \cellcolor[HTML]{E8E8E8}44.84 (\textcolor{customRed}{-\num{0.10}}) & \cellcolor[HTML]{E8E8E8}52.15 (\textcolor{customRed}{-\num{0.35}}) & \cellcolor[HTML]{E8E8E8}53.81 (\textcolor{customRed}{-\num{0.58}}) \\

\multicolumn{1}{c|}{}                        & \multicolumn{1}{c|}{ \; \ \enspace \ThickVLine -- w/I \textbf{(K=10)}}                          & \textbf{53.79} (\textcolor{darkgreen}{+\num{2.21}}) & \textbf{47.69} (\textcolor{darkgreen}{+\num{2.75}}) & \textbf{55.00} (\textcolor{darkgreen}{+\num{2.50}}) & \textbf{57.12} (\textcolor{darkgreen}{+\num{2.73}}) \\

\multicolumn{1}{c|}{}                         & \multicolumn{1}{c|}{\cellcolor[HTML]{E8E8E8} \; \enspace \ThickVLine -- w/I \textbf{(K=15)}} & \cellcolor[HTML]{E8E8E8}53.45 (\textcolor{darkgreen}{+\num{1.87}}) & \cellcolor[HTML]{E8E8E8}47.46 (\textcolor{darkgreen}{+\num{2.52}}) & \cellcolor[HTML]{E8E8E8}54.43 (\textcolor{darkgreen}{+\num{1.93}}) & \cellcolor[HTML]{E8E8E8}56.65 (\textcolor{darkgreen}{+\num{2.26}}) \\ 
\multicolumn{1}{c|}{}                        & \multicolumn{1}{l|}{\textbf{Flan-T5}}                                 & \multicolumn{4}{c}{}                                                                                                          \\
\multicolumn{1}{c|}{}                        & \multicolumn{1}{c|}{\cellcolor[HTML]{E8E8E8}  \enspace \ThickVLine -- w/I \textbf{(K=5)}}  & \cellcolor[HTML]{E8E8E8}29.11 (\textcolor{customRed}{-\num{1.65}}) & \cellcolor[HTML]{E8E8E8}24.63 (\textcolor{customRed}{-\num{1.67}}) & \cellcolor[HTML]{E8E8E8}29.86 (\textcolor{customRed}{-\num{2.0}}) & \cellcolor[HTML]{E8E8E8}31.14 (\textcolor{customRed}{-\num{2.26}}) \\

\multicolumn{1}{c|}{}                        & \multicolumn{1}{c|}{ \; \ \enspace \ThickVLine -- w/I \textbf{(K=10)}}                          & 30.74 (\textcolor{customRed}{-\num{0.02}})                         & 26.25 (\textcolor{customRed}{-\num{0.05}})                        & 31.40 (\textcolor{customRed}{-\num{0.46}})                        & 33.21 (\textcolor{customRed}{-\num{0.19}})                        \\

\multicolumn{1}{c|}{\multirow{-12}{*}{\textbf{Visual} (\textit{Image})}} & \multicolumn{1}{c|}{\cellcolor[HTML]{E8E8E8} \ \ \enspace \ThickVLine -- w/I \textbf{(K=15)}} & \cellcolor[HTML]{E8E8E8}\textbf{31.54} (\textcolor{darkgreen}{+\num{0.78}}) & \cellcolor[HTML]{E8E8E8}\textbf{26.80} (\textcolor{darkgreen}{+\num{0.5}}) & \cellcolor[HTML]{E8E8E8}\textbf{32.49} (\textcolor{darkgreen}{+\num{0.63}}) & \cellcolor[HTML]{E8E8E8}\textbf{34.24} (\textcolor{darkgreen}{+\num{0.84}}) \\ \Xhline{4\arrayrulewidth}
\end{tabular}
}
\caption{Tags Words Evaluation. \textbf{Selected Content} (Only Important Sentences). \textbf{K} denotes the number of sentences retrieved for both text and visual modalities. The best results compared to their respective baseline models are marked in \textbf{bold}, and $\Delta$ gains are shown in round brackets and highlighted with \textcolor{darkgreen}{green} and \textcolor{customRed}{red} colors.}
\label{table:tags-evaluation}
\end{table*}

\subsection{Tag Words}

\paragraph{Baselines} Table \ref{table:tags-evaluation=other} compares tag word generation across baseline models and extractive methods (\textbf{TF-IDF}, \textbf{TextRank}, and \textbf{KEA}) on an English test corpus. Extractive methods, limited to identifying tag words already present in the text, demonstrate varied performance, with TextRank lagging due to its tendency to extract longer phrases, misaligned with the brevity of reference tags. In contrast, TF-IDF and KEA perform moderately better but are outshined by abstractive models like mT5, mT0, and Flan-T5, which excel in generating concise and relevant tag words, surpassing extractive approaches that cannot synthesize new tag words. This distinction highlights the superior capability of abstractive methods in handling the tag word generation task.

\paragraph{MultiRAGen} Table \ref{table:tags-evaluation} extends the tag word evaluation results to include baselines mT5, mT0, Flan-T5, Gemini Pro, Mixtral and MultiRAGen across a multilingual text corpus. The abstractive baselines exhibit commendable performance, with mT0 demonstrating a noticeably higher gain over mT5 and Flan-T5. This improvement is attributed to the further fine-tuning of mT0 on a multitask framework, enhancing its crosslingual generalization. Although Flan-T5 produces results similar to mT0 and mT5 in English (see Table \ref{table:tags-evaluation=other}), it falls short of generating any tag words when it comes to certain languages with limited resources. This disparity is particularly noticeable in Table \ref{table:language-tags-flant5-ia}.

\subsection{Discussion}

\paragraph{Gemini Pro and Mixtral} exhibit notably low performance, indicating a tendency to generate headlines and tag words in English or  romanized rather than the native language of the article. More details with \emph{``Prompt for LLMs''} and example outputs are presented in Appendix~\ref{sec:LLM-Performance}.

\paragraph{Better content selection approach} Our experiments explored two methods: \textbf{(1)} using only top-$\mathcal{K}$ retrieved sentences and \textbf{(2)} top-$\mathcal{K}$ merging these with the original article. We found combining retrieved sentences with the full article improved headline generation (see Table \ref{table:headline-evaluation-ia} vs. \ref{table:headline-evaluation}), while using solely retrieved sentences was more effective for tag generation (see Table \ref{table:tags-evaluation} vs. \ref{table:tags-evaluation-ia}). This difference arises because tags, being more concise, benefit from focused inputs, whereas headlines may require broader context for optimal generation. 

\paragraph{High-Resource vs. Low-Resource Languages} Our model, \textbf{MultiRAGen}, achieves balanced performance across high-resource and low-resource languages. Detailed results, grouped by resource availability, are presented in the Appendix (from Table \ref{table:language-headline-mt5} to Table \ref{table:language-tags-flant5-ia}).

\section{Conclusion}

In this paper, we compile the \texttt{\textbf{XL-HeadTags}} dataset for headline and tag generation tasks, including \num{20} languages across \num{6} diverse language families. We introduce a novel content selection approach that leverages auxiliary information from both textual and visual modalities to pinpoint the most salient content within news articles. We employ instruction tuning to generate both headlines and tags in controlled and unrestricted manners. Furthermore, we have developed a suite of tools by accumulating open-source resources for processing and evaluating multilingual texts.

\section*{Limitations}
\label{sec:limitations}

\paragraph{Potential Bias in Dataset Source} Our dataset exclusively comprises articles sourced from the BBC, which may introduce a bias toward specific narratives or ideologies, potentially impacting the dataset's representativeness and diversity.

\paragraph{Handling Multiple Images and Captions} In our analysis, we noted that documents frequently contain several images and captions without a direct one-to-one correspondence. Ideally, a precise mapping between each image and its caption would enhance our retrieval process. However, due to the absence of such mappings, we treat each image and caption as separate entities for independent retrieval processes. Integrating both images and captions for simultaneous retrieval presents an opportunity for future research, potentially refining the information extraction process.

\paragraph{Computational Constraints} Given the computational constraints and resource limitations, we opted for the base versions of models for our experimentation. Despite these constraints, we introduce a novel content selection strategy that leverages auxiliary information from both textual and visual modalities. This approach aims to pinpoint the most pertinent content in news articles across multiple languages. Our designed modules are \texttt{\textbf{plug-and-play}}, ensuring they can be effortlessly integrated with language models of varying sizes. This flexibility facilitates broader applicability and enhances the adaptability of our approach in diverse computational environments.

\section*{Ethics Statement}
\label{sec:ethics}

\paragraph{Data Crawling} We took ethical consideration into account when scraping data. The data we have collected is intended exclusively for non-commercial research purposes. We conducted our web scraping activities at a reasonable rate, with no intention of causing a Distributed Denial of Service (\textbf{DDoS}) attack. Additionaly, we read the instructions listed in robots.txt\footnote{\url{https://moz.com/learn/seo/robotstxt}} of each website to ensure we were able to crawl the desired content  as per the Robots Exclusion Protocol (REP) standards\footnote{The robots.txt file is part of the robots exclusion protocol (REP), a group of web standards regulating how robots crawl the web.}.

\paragraph{Protection of Privacy} We intentionally chose not to collect certain information, including the author name, the time when the article was written, and any personal contact details such as email addresses, phone numbers, etc, for the purposes of our experiments. Consequently, our dataset does not contain any Personal Identifying Information (\textbf{PII}). This decision underscores our commitment to placing user privacy as a top priority.
\section*{Acknowledgements}

We thank all the anonymous reviewers for their valuable feedback and constructive suggestions for improving this work. Mir Tafseer Nayeem is supported by a Huawei PhD Fellowship.

\bibliography{custom}

\begin{thebibliography}{90}
\providecommand{\natexlab}[1]{#1}

\bibitem[{Aggarwal et~al.(2024)Aggarwal, Sathe, and Sitaram}]{aggarwal2024maple}
Divyanshu Aggarwal, Ashutosh Sathe, and Sunayana Sitaram. 2024.
\newblock \href {https://arxiv.org/abs/2401.07598} {Maple: Multilingual evaluation of parameter efficient finetuning of large language models}.
\newblock \emph{Preprint}, arXiv:2401.07598.

\bibitem[{Aharoni et~al.(2023)Aharoni, Narayan, Maynez, Herzig, Clark, and Lapata}]{aharoni-etal-2023-multilingual}
Roee Aharoni, Shashi Narayan, Joshua Maynez, Jonathan Herzig, Elizabeth Clark, and Mirella Lapata. 2023.
\newblock \href {https://doi.org/10.18653/v1/2023.findings-acl.220} {Multilingual summarization with factual consistency evaluation}.
\newblock In \emph{Findings of the Association for Computational Linguistics: ACL 2023}, pages 3562--3591, Toronto, Canada. Association for Computational Linguistics.

\bibitem[{Ahuja et~al.(2023)Ahuja, Diddee, Hada, Ochieng, Ramesh, Jain, Nambi, Ganu, Segal, Ahmed, Bali, and Sitaram}]{ahuja-etal-2023-mega}
Kabir Ahuja, Harshita Diddee, Rishav Hada, Millicent Ochieng, Krithika Ramesh, Prachi Jain, Akshay Nambi, Tanuja Ganu, Sameer Segal, Mohamed Ahmed, Kalika Bali, and Sunayana Sitaram. 2023.
\newblock \href {https://doi.org/10.18653/v1/2023.emnlp-main.258} {{MEGA}: Multilingual evaluation of generative {AI}}.
\newblock In \emph{Proceedings of the 2023 Conference on Empirical Methods in Natural Language Processing}, pages 4232--4267, Singapore. Association for Computational Linguistics.

\bibitem[{Akash et~al.(2023)Akash, Nayeem, Shohan, and Islam}]{akash-etal-2023-shironaam}
Abu~Ubaida Akash, Mir~Tafseer Nayeem, Faisal~Tareque Shohan, and Tanvir Islam. 2023.
\newblock \href {https://doi.org/10.18653/v1/2023.eacl-main.4} {Shironaam: {B}engali news headline generation using auxiliary information}.
\newblock In \emph{Proceedings of the 17th Conference of the European Chapter of the Association for Computational Linguistics}, pages 52--67, Dubrovnik, Croatia. Association for Computational Linguistics.

\bibitem[{Bahdanau et~al.(2014)Bahdanau, Cho, and Bengio}]{bahdanau2014neural}
Dzmitry Bahdanau, Kyunghyun Cho, and Yoshua Bengio. 2014.
\newblock \href {https://arxiv.org/abs/1409.0473} {Neural machine translation by jointly learning to align and translate}.
\newblock \emph{Preprint}, arXiv:1409.0473.

\bibitem[{Banerjee and Lavie(2005)}]{banarjee2005}
Satanjeev Banerjee and Alon Lavie. 2005.
\newblock \href {https://www.aclweb.org/anthology/W05-0909} {{METEOR}: An automatic metric for {MT} evaluation with improved correlation with human judgments}.
\newblock In \emph{Proceedings of the {ACL} Workshop on Intrinsic and Extrinsic Evaluation Measures for Machine Translation and/or Summarization}, pages 65--72, Ann Arbor, Michigan. Association for Computational Linguistics.

\bibitem[{Bommasani and Cardie(2020)}]{bommasani2020intrinsic}
Rishi Bommasani and Claire Cardie. 2020.
\newblock \href {https://doi.org/10.18653/v1/2020.emnlp-main.649} {Intrinsic evaluation of summarization datasets}.
\newblock In \emph{Proceedings of the 2020 Conference on Empirical Methods in Natural Language Processing (EMNLP)}, pages 8075--8096, Online. Association for Computational Linguistics.

\bibitem[{Borgeaud et~al.(2021)Borgeaud, Mensch, Hoffmann, Cai, Rutherford, Millican, Driessche, Lespiau, Damoc, Clark et~al.}]{borgeaud2021improving}
Sebastian Borgeaud, Arthur Mensch, Jordan Hoffmann, Trevor Cai, Eliza Rutherford, Katie Millican, George van~den Driessche, Jean-Baptiste Lespiau, Bogdan Damoc, Aidan Clark, et~al. 2021.
\newblock Improving language models by retrieving from trillions of tokens.
\newblock \emph{arXiv preprint arXiv:2112.04426}.

\bibitem[{Boudin(2016)}]{boudin:2016:COLINGDEMO}
Florian Boudin. 2016.
\newblock \href {http://aclweb.org/anthology/C16-2015} {pke: an open source python-based keyphrase extraction toolkit}.
\newblock In \emph{Proceedings of COLING 2016, the 26th International Conference on Computational Linguistics: System Demonstrations}, pages 69--73, Osaka, Japan.

\bibitem[{Brown et~al.(2020)Brown, Mann, Ryder, Subbiah, Kaplan, Dhariwal, Neelakantan, Shyam, Sastry, Askell et~al.}]{brown2020language}
Tom Brown, Benjamin Mann, Nick Ryder, Melanie Subbiah, Jared~D Kaplan, Prafulla Dhariwal, Arvind Neelakantan, Pranav Shyam, Girish Sastry, Amanda Askell, et~al. 2020.
\newblock Language models are few-shot learners.
\newblock \emph{Advances in neural information processing systems}, 33:1877--1901.

\bibitem[{Bukhtiyarov and Gusev(2020)}]{Transformer-Based-Models-for-News-Headline-Generation}
Alexey Bukhtiyarov and Ilya Gusev. 2020.
\newblock Advances of transformer-based models for news headline generation.
\newblock In \emph{Artificial Intelligence and Natural Language}, pages 54--61, Cham. Springer International Publishing.

\bibitem[{Cai et~al.(2019)Cai, Wang, Bi, Tu, Liu, Lam, and Shi}]{cai-etal-2019-skeleton}
Deng Cai, Yan Wang, Wei Bi, Zhaopeng Tu, Xiaojiang Liu, Wai Lam, and Shuming Shi. 2019.
\newblock \href {https://doi.org/10.18653/v1/N19-1124} {Skeleton-to-response: Dialogue generation guided by retrieval memory}.
\newblock In \emph{Proceedings of the 2019 Conference of the North {A}merican Chapter of the Association for Computational Linguistics: Human Language Technologies, Volume 1 (Long and Short Papers)}, pages 1219--1228, Minneapolis, Minnesota. Association for Computational Linguistics.

\bibitem[{Cao et~al.(2017)Cao, Li, Li, and Wei}]{10.5555/3298483.3298678}
Ziqiang Cao, Wenjie Li, Sujian Li, and Furu Wei. 2017.
\newblock Improving multi-document summarization via text classification.
\newblock In \emph{Proceedings of the Thirty-First AAAI Conference on Artificial Intelligence}, AAAI'17, page 3053–3059. AAAI Press.

\bibitem[{Chali et~al.(2017)Chali, Tanvee, and Nayeem}]{chali-etal-2017-towards}
Yllias Chali, Moin Tanvee, and Mir~Tafseer Nayeem. 2017.
\newblock \href {https://aclanthology.org/I17-2071} {Towards abstractive multi-document summarization using submodular function-based framework, sentence compression and merging}.
\newblock In \emph{Proceedings of the Eighth International Joint Conference on Natural Language Processing (Volume 2: Short Papers)}, pages 418--424, Taipei, Taiwan. Asian Federation of Natural Language Processing.

\bibitem[{Chan et~al.(2019)Chan, Chen, Wang, and King}]{chan-etal-2019-neural}
Hou~Pong Chan, Wang Chen, Lu~Wang, and Irwin King. 2019.
\newblock \href {https://doi.org/10.18653/v1/P19-1208} {Neural keyphrase generation via reinforcement learning with adaptive rewards}.
\newblock In \emph{Proceedings of the 57th Annual Meeting of the Association for Computational Linguistics}, pages 2163--2174, Florence, Italy. Association for Computational Linguistics.

\bibitem[{Chen and Zhuge(2018)}]{chen-zhuge-2018-abstractive}
Jingqiang Chen and Hai Zhuge. 2018.
\newblock \href {https://doi.org/10.18653/v1/D18-1438} {Abstractive text-image summarization using multi-modal attentional hierarchical {RNN}}.
\newblock In \emph{Proceedings of the 2018 Conference on Empirical Methods in Natural Language Processing}, pages 4046--4056, Brussels, Belgium. Association for Computational Linguistics.

\bibitem[{Chen et~al.(2020)Chen, Chan, Li, and King}]{chen-etal-2020-exclusive}
Wang Chen, Hou~Pong Chan, Piji Li, and Irwin King. 2020.
\newblock \href {https://doi.org/10.18653/v1/2020.acl-main.103} {Exclusive hierarchical decoding for deep keyphrase generation}.
\newblock In \emph{Proceedings of the 58th Annual Meeting of the Association for Computational Linguistics}, pages 1095--1105, Online. Association for Computational Linguistics.

\bibitem[{Chen et~al.(2022)Chen, Hu, Chen, Verga, and Cohen}]{chen-etal-2022-murag}
Wenhu Chen, Hexiang Hu, Xi~Chen, Pat Verga, and William Cohen. 2022.
\newblock \href {https://doi.org/10.18653/v1/2022.emnlp-main.375} {{M}u{RAG}: Multimodal retrieval-augmented generator for open question answering over images and text}.
\newblock In \emph{Proceedings of the 2022 Conference on Empirical Methods in Natural Language Processing}, pages 5558--5570, Abu Dhabi, United Arab Emirates. Association for Computational Linguistics.

\bibitem[{Chowdhury et~al.(2021)Chowdhury, Nayeem, Mim, Chowdhury, and Jannat}]{chowdhury-etal-2021-unsupervised}
Radia~Rayan Chowdhury, Mir~Tafseer Nayeem, Tahsin~Tasnim Mim, Md. Saifur~Rahman Chowdhury, and Taufiqul Jannat. 2021.
\newblock \href {https://doi.org/10.18653/v1/2021.eacl-main.224} {Unsupervised abstractive summarization of {B}engali text documents}.
\newblock In \emph{Proceedings of the 16th Conference of the European Chapter of the Association for Computational Linguistics: Main Volume}, pages 2612--2619, Online. Association for Computational Linguistics.

\bibitem[{Chronopoulou et~al.(2023)Chronopoulou, Pfeiffer, Maynez, Wang, Ruder, and Agrawal}]{chronopoulou2023language}
Alexandra Chronopoulou, Jonas Pfeiffer, Joshua Maynez, Xinyi Wang, Sebastian Ruder, and Priyanka Agrawal. 2023.
\newblock \href {https://arxiv.org/abs/2311.09344} {Language and task arithmetic with parameter-efficient layers for zero-shot summarization}.
\newblock \emph{Preprint}, arXiv:2311.09344.

\bibitem[{Chung et~al.(2022)Chung, Hou, Longpre, Zoph, Tay, Fedus, Li, Wang, Dehghani, Brahma, Webson, Gu, Dai, Suzgun, Chen, Chowdhery, Castro-Ros, Pellat, Robinson, Valter, Narang, Mishra, Yu, Zhao, Huang, Dai, Yu, Petrov, Chi, Dean, Devlin, Roberts, Zhou, Le, and Wei}]{chung2022scaling}
Hyung~Won Chung, Le~Hou, Shayne Longpre, Barret Zoph, Yi~Tay, William Fedus, Yunxuan Li, Xuezhi Wang, Mostafa Dehghani, Siddhartha Brahma, Albert Webson, Shixiang~Shane Gu, Zhuyun Dai, Mirac Suzgun, Xinyun Chen, Aakanksha Chowdhery, Alex Castro-Ros, Marie Pellat, Kevin Robinson, Dasha Valter, Sharan Narang, Gaurav Mishra, Adams Yu, Vincent Zhao, Yanping Huang, Andrew Dai, Hongkun Yu, Slav Petrov, Ed~H. Chi, Jeff Dean, Jacob Devlin, Adam Roberts, Denny Zhou, Quoc~V. Le, and Jason Wei. 2022.
\newblock \href {https://arxiv.org/abs/2210.11416} {Scaling instruction-finetuned language models}.
\newblock \emph{Preprint}, arXiv:2210.11416.

\bibitem[{Conneau et~al.(2020)Conneau, Khandelwal, Goyal, Chaudhary, Wenzek, Guzm{\'a}n, Grave, Ott, Zettlemoyer, and Stoyanov}]{conneau-etal-2020-unsupervised}
Alexis Conneau, Kartikay Khandelwal, Naman Goyal, Vishrav Chaudhary, Guillaume Wenzek, Francisco Guzm{\'a}n, Edouard Grave, Myle Ott, Luke Zettlemoyer, and Veselin Stoyanov. 2020.
\newblock \href {https://doi.org/10.18653/v1/2020.acl-main.747} {Unsupervised cross-lingual representation learning at scale}.
\newblock In \emph{Proceedings of the 58th Annual Meeting of the Association for Computational Linguistics}, pages 8440--8451, Online. Association for Computational Linguistics.

\bibitem[{Devlin et~al.(2019)Devlin, Chang, Lee, and Toutanova}]{devlin2019bert}
Jacob Devlin, Ming-Wei Chang, Kenton Lee, and Kristina Toutanova. 2019.
\newblock Bert: Pre-training of deep bidirectional transformers for language understanding.
\newblock In \emph{Proceedings of the 2019 Conference of the North American Chapter of the Association for Computational Linguistics: Human Language Technologies, Volume 1 (Long and Short Papers)}, pages 4171--4186.

\bibitem[{Dong et~al.(2015)Dong, Wu, He, Yu, and Wang}]{dong-etal-2015-multi}
Daxiang Dong, Hua Wu, Wei He, Dianhai Yu, and Haifeng Wang. 2015.
\newblock \href {https://doi.org/10.3115/v1/P15-1166} {Multi-task learning for multiple language translation}.
\newblock In \emph{Proceedings of the 53rd Annual Meeting of the Association for Computational Linguistics and the 7th International Joint Conference on Natural Language Processing (Volume 1: Long Papers)}, pages 1723--1732, Beijing, China. Association for Computational Linguistics.

\bibitem[{Fuad et~al.(2019)Fuad, Nayeem, Mahmud, and Chali}]{FUAD2019216}
Tanvir~Ahmed Fuad, Mir~Tafseer Nayeem, Asif Mahmud, and Yllias Chali. 2019.
\newblock \href {https://doi.org/10.1016/j.csl.2019.04.006} {Neural sentence fusion for diversity driven abstractive multi-document summarization}.
\newblock \emph{Computer Speech \& Language}, 58:216--230.

\bibitem[{Gu et~al.(2016)Gu, Lu, Li, and Li}]{gu-etal-2016-incorporating}
Jiatao Gu, Zhengdong Lu, Hang Li, and Victor~O.K. Li. 2016.
\newblock \href {https://doi.org/10.18653/v1/P16-1154} {Incorporating copying mechanism in sequence-to-sequence learning}.
\newblock In \emph{Proceedings of the 54th Annual Meeting of the Association for Computational Linguistics (Volume 1: Long Papers)}, pages 1631--1640, Berlin, Germany. Association for Computational Linguistics.

\bibitem[{Gu et~al.(2020)Gu, Mao, Han, Liu, Wu, Yu, Finnie, Yu, Zhai, and Zukoski}]{10.1145/3366423.3380247}
Xiaotao Gu, Yuning Mao, Jiawei Han, Jialu Liu, You Wu, Cong Yu, Daniel Finnie, Hongkun Yu, Jiaqi Zhai, and Nicholas Zukoski. 2020.
\newblock \href {https://doi.org/10.1145/3366423.3380247} {Generating representative headlines for news stories}.
\newblock In \emph{Proceedings of The Web Conference 2020}, WWW '20, page 1773–1784, New York, NY, USA. Association for Computing Machinery.

\bibitem[{Guo et~al.(2018)Guo, Pasunuru, and Bansal}]{guo-etal-2018-soft}
Han Guo, Ramakanth Pasunuru, and Mohit Bansal. 2018.
\newblock \href {https://doi.org/10.18653/v1/P18-1064} {Soft layer-specific multi-task summarization with entailment and question generation}.
\newblock In \emph{Proceedings of the 56th Annual Meeting of the Association for Computational Linguistics (Volume 1: Long Papers)}, pages 687--697, Melbourne, Australia. Association for Computational Linguistics.

\bibitem[{Guu et~al.(2020)Guu, Lee, Tung, Pasupat, and Chang}]{pmlr-v119-guu20a}
Kelvin Guu, Kenton Lee, Zora Tung, Panupong Pasupat, and Mingwei Chang. 2020.
\newblock Retrieval augmented language model pre-training.
\newblock In \emph{Proceedings of the 37th International Conference on Machine Learning}, volume 119 of \emph{Proceedings of Machine Learning Research}, pages 3929--3938. PMLR.

\bibitem[{Hasan et~al.(2021)Hasan, Bhattacharjee, Islam, Mubasshir, Li, Kang, Rahman, and Shahriyar}]{hasan-etal-2021-xl}
Tahmid Hasan, Abhik Bhattacharjee, Md.~Saiful Islam, Kazi Mubasshir, Yuan-Fang Li, Yong-Bin Kang, M.~Sohel Rahman, and Rifat Shahriyar. 2021.
\newblock \href {https://doi.org/10.18653/v1/2021.findings-acl.413} {{XL}-sum: Large-scale multilingual abstractive summarization for 44 languages}.
\newblock In \emph{Findings of the Association for Computational Linguistics: ACL-IJCNLP 2021}, pages 4693--4703, Online. Association for Computational Linguistics.

\bibitem[{He et~al.(2021)He, Huang, Cui, Li, and Liu}]{he-etal-2021-fast}
Qiuxiang He, Guoping Huang, Qu~Cui, Li~Li, and Lemao Liu. 2021.
\newblock \href {https://doi.org/10.18653/v1/2021.acl-long.246} {Fast and accurate neural machine translation with translation memory}.
\newblock In \emph{Proceedings of the 59th Annual Meeting of the Association for Computational Linguistics and the 11th International Joint Conference on Natural Language Processing (Volume 1: Long Papers)}, pages 3170--3180, Online. Association for Computational Linguistics.

\bibitem[{Higurashi et~al.(2018)Higurashi, Kobayashi, Masuyama, and Murao}]{higurashi-etal-2018-extractive}
Tatsuru Higurashi, Hayato Kobayashi, Takeshi Masuyama, and Kazuma Murao. 2018.
\newblock \href {https://aclanthology.org/C18-1148} {Extractive headline generation based on learning to rank for community question answering}.
\newblock In \emph{Proceedings of the 27th International Conference on Computational Linguistics}, pages 1742--1753, Santa Fe, New Mexico, USA. Association for Computational Linguistics.

\bibitem[{Iwama and Kano(2019)}]{iwama-kano-2019-multiple}
Kango Iwama and Yoshinobu Kano. 2019.
\newblock \href {https://doi.org/10.18653/v1/W19-8612} {Multiple news headlines generation using page metadata}.
\newblock In \emph{Proceedings of the 12th International Conference on Natural Language Generation}, pages 101--105, Tokyo, Japan. Association for Computational Linguistics.

\bibitem[{Jiang et~al.(2023)Jiang, Sablayrolles, Mensch, Bamford, Chaplot, de~las Casas, Bressand, Lengyel, Lample, Saulnier, Lavaud, Lachaux, Stock, Scao, Lavril, Wang, Lacroix, and Sayed}]{jiang2023mistral}
Albert~Q. Jiang, Alexandre Sablayrolles, Arthur Mensch, Chris Bamford, Devendra~Singh Chaplot, Diego de~las Casas, Florian Bressand, Gianna Lengyel, Guillaume Lample, Lucile Saulnier, Lélio~Renard Lavaud, Marie-Anne Lachaux, Pierre Stock, Teven~Le Scao, Thibaut Lavril, Thomas Wang, Timothée Lacroix, and William~El Sayed. 2023.
\newblock \href {https://arxiv.org/abs/2310.06825} {Mistral 7b}.
\newblock \emph{Preprint}, arXiv:2310.06825.

\bibitem[{Khandelwal et~al.(2020)Khandelwal, Levy, Jurafsky, Zettlemoyer, and Lewis}]{khandelwal2020generalization}
Urvashi Khandelwal, Omer Levy, Dan Jurafsky, Luke Zettlemoyer, and Mike Lewis. 2020.
\newblock \href {https://arxiv.org/abs/1911.00172} {Generalization through memorization: Nearest neighbor language models}.
\newblock \emph{Preprint}, arXiv:1911.00172.

\bibitem[{Krubi{\'n}ski and Pecina(2023)}]{krubinski-pecina-2023-mlask}
Mateusz Krubi{\'n}ski and Pavel Pecina. 2023.
\newblock \href {https://doi.org/10.18653/v1/2023.findings-eacl.67} {{MLASK}: Multimodal summarization of video-based news articles}.
\newblock In \emph{Findings of the Association for Computational Linguistics: EACL 2023}, pages 910--924, Dubrovnik, Croatia. Association for Computational Linguistics.

\bibitem[{Kumar et~al.(2022)Kumar, Shrotriya, Sahu, Dabre, Puduppully, Kunchukuttan, Mishra, Khapra, and Kumar}]{kumar2022indicnlg}
Aman Kumar, Himani Shrotriya, Prachi Sahu, Raj Dabre, Ratish Puduppully, Anoop Kunchukuttan, Amogh Mishra, Mitesh~M Khapra, and Pratyush Kumar. 2022.
\newblock \href {https://arxiv.org/abs/2203.05437} {Indicnlg suite: Multilingual datasets for diverse nlg tasks in indic languages}.
\newblock \emph{arXiv preprint arXiv:2203.05437}.

\bibitem[{Lewis et~al.(2020{\natexlab{a}})Lewis, Perez, Piktus, Petroni, Karpukhin, Goyal, K\"{u}ttler, Lewis, Yih, Rockt\"{a}schel, Riedel, and Kiela}]{10.5555/3495724.3496517}
Patrick Lewis, Ethan Perez, Aleksandra Piktus, Fabio Petroni, Vladimir Karpukhin, Naman Goyal, Heinrich K\"{u}ttler, Mike Lewis, Wen-tau Yih, Tim Rockt\"{a}schel, Sebastian Riedel, and Douwe Kiela. 2020{\natexlab{a}}.
\newblock Retrieval-augmented generation for knowledge-intensive nlp tasks.
\newblock In \emph{Proceedings of the 34th International Conference on Neural Information Processing Systems}, NIPS'20, Red Hook, NY, USA. Curran Associates Inc.

\bibitem[{Lewis et~al.(2020{\natexlab{b}})Lewis, Perez, Piktus, Petroni, Karpukhin, Goyal, K{\"u}ttler, Lewis, Yih, Rockt{\"a}schel et~al.}]{lewis2020retrieval}
Patrick Lewis, Ethan Perez, Aleksandra Piktus, Fabio Petroni, Vladimir Karpukhin, Naman Goyal, Heinrich K{\"u}ttler, Mike Lewis, Wen-tau Yih, Tim Rockt{\"a}schel, et~al. 2020{\natexlab{b}}.
\newblock Retrieval-augmented generation for knowledge-intensive nlp tasks.
\newblock \emph{Advances in Neural Information Processing Systems}, 33:9459--9474.

\bibitem[{Li et~al.(2018)Li, Zhu, Liu, Zhang, and Zong}]{ijcai2018p577}
Haoran Li, Junnan Zhu, Tianshang Liu, Jiajun Zhang, and Chengqing Zong. 2018.
\newblock \href {https://doi.org/10.24963/ijcai.2018/577} {Multi-modal sentence summarization with modality attention and image filtering}.
\newblock In \emph{Proceedings of the Twenty-Seventh International Joint Conference on Artificial Intelligence, {IJCAI-18}}, pages 4152--4158. International Joint Conferences on Artificial Intelligence Organization.

\bibitem[{Lin(2004)}]{lin-2004-rouge}
Chin-Yew Lin. 2004.
\newblock \href {https://www.aclweb.org/anthology/W04-1013} {{ROUGE}: A package for automatic evaluation of summaries}.
\newblock In \emph{Text Summarization Branches Out}, pages 74--81, Barcelona, Spain. Association for Computational Linguistics.

\bibitem[{Liu et~al.(2021)Liu, Wang, Wang, and Ordonez}]{liu-etal-2021-visual}
Fuxiao Liu, Yinghan Wang, Tianlu Wang, and Vicente Ordonez. 2021.
\newblock \href {https://doi.org/10.18653/v1/2021.emnlp-main.542} {Visual news: Benchmark and challenges in news image captioning}.
\newblock In \emph{Proceedings of the 2021 Conference on Empirical Methods in Natural Language Processing}, pages 6761--6771, Online and Punta Cana, Dominican Republic. Association for Computational Linguistics.

\bibitem[{Liu et~al.(2023)Liu, Lin, Hewitt, Paranjape, Bevilacqua, Petroni, and Liang}]{liu2023lost}
Nelson~F. Liu, Kevin Lin, John Hewitt, Ashwin Paranjape, Michele Bevilacqua, Fabio Petroni, and Percy Liang. 2023.
\newblock \href {https://arxiv.org/abs/2307.03172} {Lost in the middle: How language models use long contexts}.
\newblock \emph{Preprint}, arXiv:2307.03172.

\bibitem[{Loshchilov and Hutter(2019)}]{adam-w-loshchilov2019decoupled}
Ilya Loshchilov and Frank Hutter. 2019.
\newblock \href {https://arxiv.org/abs/1711.05101} {Decoupled weight decay regularization}.
\newblock \emph{Preprint}, arXiv:1711.05101.

\bibitem[{Meng et~al.(2017)Meng, Zhao, Han, He, Brusilovsky, and Chi}]{meng-etal-2017-deep}
Rui Meng, Sanqiang Zhao, Shuguang Han, Daqing He, Peter Brusilovsky, and Yu~Chi. 2017.
\newblock \href {https://doi.org/10.18653/v1/P17-1054} {Deep keyphrase generation}.
\newblock In \emph{Proceedings of the 55th Annual Meeting of the Association for Computational Linguistics (Volume 1: Long Papers)}, pages 582--592, Vancouver, Canada. Association for Computational Linguistics.

\bibitem[{Mihalcea and Tarau(2004)}]{mihalcea-tarau-2004-textrank}
Rada Mihalcea and Paul Tarau. 2004.
\newblock \href {https://aclanthology.org/W04-3252} {{T}ext{R}ank: Bringing order into text}.
\newblock In \emph{Proceedings of the 2004 Conference on Empirical Methods in Natural Language Processing}, pages 404--411, Barcelona, Spain. Association for Computational Linguistics.

\bibitem[{Min et~al.(2022)Min, Lewis, Zettlemoyer, and Hajishirzi}]{min-etal-2022-metaicl}
Sewon Min, Mike Lewis, Luke Zettlemoyer, and Hannaneh Hajishirzi. 2022.
\newblock \href {https://doi.org/10.18653/v1/2022.naacl-main.201} {{M}eta{ICL}: Learning to learn in context}.
\newblock In \emph{Proceedings of the 2022 Conference of the North American Chapter of the Association for Computational Linguistics: Human Language Technologies}, pages 2791--2809, Seattle, United States. Association for Computational Linguistics.

\bibitem[{Muennighoff et~al.(2022)Muennighoff, Wang, Sutawika, Roberts, Biderman, Scao, Bari, Shen, Yong, Schoelkopf et~al.}]{mt0-muennighoff2022crosslingual}
Niklas Muennighoff, Thomas Wang, Lintang Sutawika, Adam Roberts, Stella Biderman, Teven~Le Scao, M~Saiful Bari, Sheng Shen, Zheng-Xin Yong, Hailey Schoelkopf, et~al. 2022.
\newblock Crosslingual generalization through multitask finetuning.
\newblock \emph{arXiv preprint arXiv:2211.01786}.

\bibitem[{Narayan et~al.(2018{\natexlab{a}})Narayan, Cohen, and Lapata}]{narayan-etal-2018-dont}
Shashi Narayan, Shay~B. Cohen, and Mirella Lapata. 2018{\natexlab{a}}.
\newblock \href {https://doi.org/10.18653/v1/D18-1206} {Don{'}t give me the details, just the summary! topic-aware convolutional neural networks for extreme summarization}.
\newblock In \emph{Proceedings of the 2018 Conference on Empirical Methods in Natural Language Processing}, pages 1797--1807, Brussels, Belgium. Association for Computational Linguistics.

\bibitem[{Narayan et~al.(2018{\natexlab{b}})Narayan, Cohen, and Lapata}]{narayan2018don}
Shashi Narayan, Shay~B. Cohen, and Mirella Lapata. 2018{\natexlab{b}}.
\newblock \href {https://doi.org/10.18653/v1/D18-1206} {Don{'}t give me the details, just the summary! topic-aware convolutional neural networks for extreme summarization}.
\newblock In \emph{Proceedings of the 2018 Conference on Empirical Methods in Natural Language Processing}, pages 1797--1807, Brussels, Belgium. Association for Computational Linguistics.

\bibitem[{Nayeem and Chali(2017{\natexlab{a}})}]{nayeem-chali-2017-extract}
Mir~Tafseer Nayeem and Yllias Chali. 2017{\natexlab{a}}.
\newblock \href {https://doi.org/10.18653/v1/W17-2407} {Extract with order for coherent multi-document summarization}.
\newblock In \emph{Proceedings of {T}ext{G}raphs-11: the Workshop on Graph-based Methods for Natural Language Processing}, pages 51--56, Vancouver, Canada. Association for Computational Linguistics.

\bibitem[{Nayeem and Chali(2017{\natexlab{b}})}]{10.1145/3132847.3133106}
Mir~Tafseer Nayeem and Yllias Chali. 2017{\natexlab{b}}.
\newblock \href {https://doi.org/10.1145/3132847.3133106} {Paraphrastic fusion for abstractive multi-sentence compression generation}.
\newblock In \emph{Proceedings of the 2017 ACM on Conference on Information and Knowledge Management}, CIKM '17, page 2223–2226, New York, NY, USA. Association for Computing Machinery.

\bibitem[{Nayeem et~al.(2018)Nayeem, Fuad, and Chali}]{nayeem-etal-2018-abstractive}
Mir~Tafseer Nayeem, Tanvir~Ahmed Fuad, and Yllias Chali. 2018.
\newblock \href {https://aclanthology.org/C18-1102} {Abstractive unsupervised multi-document summarization using paraphrastic sentence fusion}.
\newblock In \emph{Proceedings of the 27th International Conference on Computational Linguistics}, pages 1191--1204, Santa Fe, New Mexico, USA. Association for Computational Linguistics.

\bibitem[{Nayeem et~al.(2019)Nayeem, Fuad, and Chali}]{nayeem-etal-2019-diverse}
Mir~Tafseer Nayeem, Tanvir~Ahmed Fuad, and Yllias Chali. 2019.
\newblock \href {https://doi.org/10.1007/978-3-030-15719-7_14} {Neural diverse abstractive sentence compression generation}.
\newblock In \emph{Advances in Information Retrieval}, pages 109--116, Cham. Springer International Publishing.

\bibitem[{Oostdijk et~al.(2020)Oostdijk, van Halteren, Ba{\textcommabelow{s}}ar, and Larson}]{oostdijk-etal-2020-connection}
Nelleke Oostdijk, Hans van Halteren, Erkan Ba{\textcommabelow{s}}ar, and Martha Larson. 2020.
\newblock \href {https://aclanthology.org/2020.lrec-1.535} {The connection between the text and images of news articles: New insights for multimedia analysis}.
\newblock In \emph{Proceedings of the Twelfth Language Resources and Evaluation Conference}, pages 4343--4351, Marseille, France. European Language Resources Association.

\bibitem[{Ouyang et~al.(2022)Ouyang, Wu, Jiang, Almeida, Wainwright, Mishkin, Zhang, Agarwal, Slama, Ray, Schulman, Hilton, Kelton, Miller, Simens, Askell, Welinder, Christiano, Leike, and Lowe}]{NEURIPS2022-b1efde53}
Long Ouyang, Jeffrey Wu, Xu~Jiang, Diogo Almeida, Carroll Wainwright, Pamela Mishkin, Chong Zhang, Sandhini Agarwal, Katarina Slama, Alex Ray, John Schulman, Jacob Hilton, Fraser Kelton, Luke Miller, Maddie Simens, Amanda Askell, Peter Welinder, Paul~F Christiano, Jan Leike, and Ryan Lowe. 2022.
\newblock \href {https://proceedings.neurips.cc/paper_files/paper/2022/file/b1efde53be364a73914f58805a001731-Paper-Conference.pdf} {Training language models to follow instructions with human feedback}.
\newblock In \emph{Advances in Neural Information Processing Systems}, volume~35, pages 27730--27744. Curran Associates, Inc.

\bibitem[{Papineni et~al.(2002)Papineni, Roukos, Ward, and Zhu}]{Papineni02bleu:a}
Kishore Papineni, Salim Roukos, Todd Ward, and Wei-Jing Zhu. 2002.
\newblock \href {https://doi.org/10.3115/1073083.1073135} {{B}leu: a method for automatic evaluation of machine translation}.
\newblock In \emph{Proceedings of the 40th Annual Meeting of the Association for Computational Linguistics}, pages 311--318, Philadelphia, Pennsylvania, USA. Association for Computational Linguistics.

\bibitem[{Peng et~al.(2019)Peng, Parikh, Faruqui, Dhingra, and Das}]{peng-etal-2019-text}
Hao Peng, Ankur Parikh, Manaal Faruqui, Bhuwan Dhingra, and Dipanjan Das. 2019.
\newblock \href {https://doi.org/10.18653/v1/N19-1263} {Text generation with exemplar-based adaptive decoding}.
\newblock In \emph{Proceedings of the 2019 Conference of the North {A}merican Chapter of the Association for Computational Linguistics: Human Language Technologies, Volume 1 (Long and Short Papers)}, pages 2555--2565, Minneapolis, Minnesota. Association for Computational Linguistics.

\bibitem[{Radford et~al.(2021)Radford, Kim, Hallacy, Ramesh, Goh, Agarwal, Sastry, Askell, Mishkin, Clark, Krueger, and Sutskever}]{pmlr-v139-radford21a}
Alec Radford, Jong~Wook Kim, Chris Hallacy, Aditya Ramesh, Gabriel Goh, Sandhini Agarwal, Girish Sastry, Amanda Askell, Pamela Mishkin, Jack Clark, Gretchen Krueger, and Ilya Sutskever. 2021.
\newblock \href {https://proceedings.mlr.press/v139/radford21a.html} {Learning transferable visual models from natural language supervision}.
\newblock In \emph{Proceedings of the 38th International Conference on Machine Learning}, volume 139 of \emph{Proceedings of Machine Learning Research}, pages 8748--8763. PMLR.

\bibitem[{Raffel et~al.(2020)Raffel, Shazeer, Roberts, Lee, Narang, Matena, Zhou, Li, and Liu}]{raffel2020t5}
Colin Raffel, Noam Shazeer, Adam Roberts, Katherine Lee, Sharan Narang, Michael Matena, Yanqi Zhou, Wei Li, and Peter~J Liu. 2020.
\newblock Exploring the limits of transfer learning with a unified text-to-text transformer.
\newblock In \emph{Journal of Machine Learning Research}.

\bibitem[{Ravaut et~al.(2024)Ravaut, Sun, Chen, and Joty}]{ravaut2024context}
Mathieu Ravaut, Aixin Sun, Nancy~F. Chen, and Shafiq Joty. 2024.
\newblock \href {https://arxiv.org/abs/2310.10570} {On context utilization in summarization with large language models}.
\newblock \emph{Preprint}, arXiv:2310.10570.

\bibitem[{Reimers and Gurevych(2019)}]{reimers-2019-sentence-bert}
Nils Reimers and Iryna Gurevych. 2019.
\newblock \href {https://arxiv.org/abs/1908.10084} {Sentence-bert: Sentence embeddings using siamese bert-networks}.
\newblock In \emph{Proceedings of the 2019 Conference on Empirical Methods in Natural Language Processing}. Association for Computational Linguistics.

\bibitem[{Rush et~al.(2015)Rush, Chopra, and Weston}]{rush2015neural}
Alexander~M. Rush, Sumit Chopra, and Jason Weston. 2015.
\newblock \href {https://doi.org/10.18653/v1/D15-1044} {A neural attention model for abstractive sentence summarization}.
\newblock In \emph{Proceedings of the 2015 Conference on Empirical Methods in Natural Language Processing}, pages 379--389, Lisbon, Portugal. Association for Computational Linguistics.

\bibitem[{Sanh et~al.(2022)Sanh, Webson, Raffel, Bach, Sutawika, Alyafeai, Chaffin, Stiegler, Raja, Dey, Bari, Xu, Thakker, Sharma, Szczechla, Kim, Chhablani, Nayak, Datta, Chang, Jiang, Wang, Manica, Shen, Yong, Pandey, Bawden, Wang, Neeraj, Rozen, Sharma, Santilli, Fevry, Fries, Teehan, Scao, Biderman, Gao, Wolf, and Rush}]{sanh2022multitask}
Victor Sanh, Albert Webson, Colin Raffel, Stephen Bach, Lintang Sutawika, Zaid Alyafeai, Antoine Chaffin, Arnaud Stiegler, Arun Raja, Manan Dey, M~Saiful Bari, Canwen Xu, Urmish Thakker, Shanya~Sharma Sharma, Eliza Szczechla, Taewoon Kim, Gunjan Chhablani, Nihal Nayak, Debajyoti Datta, Jonathan Chang, Mike Tian-Jian Jiang, Han Wang, Matteo Manica, Sheng Shen, Zheng~Xin Yong, Harshit Pandey, Rachel Bawden, Thomas Wang, Trishala Neeraj, Jos Rozen, Abheesht Sharma, Andrea Santilli, Thibault Fevry, Jason~Alan Fries, Ryan Teehan, Teven~Le Scao, Stella Biderman, Leo Gao, Thomas Wolf, and Alexander~M Rush. 2022.
\newblock \href {https://openreview.net/forum?id=9Vrb9D0WI4} {Multitask prompted training enables zero-shot task generalization}.
\newblock In \emph{International Conference on Learning Representations}.

\bibitem[{Scialom et~al.(2020)Scialom, Dray, Lamprier, Piwowarski, and Staiano}]{scialom-etal-2020-mlsum}
Thomas Scialom, Paul-Alexis Dray, Sylvain Lamprier, Benjamin Piwowarski, and Jacopo Staiano. 2020.
\newblock \href {https://doi.org/10.18653/v1/2020.emnlp-main.647} {{MLSUM}: The multilingual summarization corpus}.
\newblock In \emph{Proceedings of the 2020 Conference on Empirical Methods in Natural Language Processing (EMNLP)}, pages 8051--8067, Online. Association for Computational Linguistics.

\bibitem[{See et~al.(2017)See, Liu, and Manning}]{see-etal-2017-get}
Abigail See, Peter~J. Liu, and Christopher~D. Manning. 2017.
\newblock \href {https://doi.org/10.18653/v1/P17-1099} {Get to the point: Summarization with pointer-generator networks}.
\newblock In \emph{Proceedings of the 55th Annual Meeting of the Association for Computational Linguistics (Volume 1: Long Papers)}, pages 1073--1083, Vancouver, Canada. Association for Computational Linguistics.

\bibitem[{Song et~al.(2020)Song, Shuai, Yeh, Wu, Ku, and Peng}]{Song_Shuai_Yeh_Wu_Ku_Peng_2020_Attractive_or_Faithful}
Yun-Zhu Song, Hong-Han Shuai, Sung-Lin Yeh, Yi-Lun Wu, Lun-Wei Ku, and Wen-Chih Peng. 2020.
\newblock \href {https://doi.org/10.1609/aaai.v34i05.6421} {Attractive or faithful? popularity-reinforced learning for inspired headline generation}.
\newblock \emph{Proceedings of the AAAI Conference on Artificial Intelligence}, 34(05):8910--8917.

\bibitem[{Sparck~Jones(1988)}]{tfidf}
Karen Sparck~Jones. 1988.
\newblock \emph{A Statistical Interpretation of Term Specificity and Its Application in Retrieval}, page 132–142.
\newblock Taylor Graham Publishing, GBR.

\bibitem[{Sweller(2011)}]{SWELLER201137}
John Sweller. 2011.
\newblock \href {https://doi.org/10.1016/B978-0-12-387691-1.00002-8} {Chapter two - cognitive load theory}.
\newblock volume~55 of \emph{Psychology of Learning and Motivation}, pages 37--76. Academic Press.

\bibitem[{Takase et~al.(2016)Takase, Suzuki, Okazaki, Hirao, and Nagata}]{takase2016neural}
Sho Takase, Jun Suzuki, Naoaki Okazaki, Tsutomu Hirao, and Masaaki Nagata. 2016.
\newblock \href {https://doi.org/10.18653/v1/D16-1112} {Neural headline generation on {A}bstract {M}eaning {R}epresentation}.
\newblock In \emph{Proceedings of the 2016 Conference on Empirical Methods in Natural Language Processing}, pages 1054--1059, Austin, Texas. Association for Computational Linguistics.

\bibitem[{Tan et~al.(2017)Tan, Wan, and Xiao}]{tan2017neural}
Jiwei Tan, Xiaojun Wan, and Jianguo Xiao. 2017.
\newblock \href {https://dl.acm.org/doi/abs/10.5555/3171837.3171860} {From neural sentence summarization to headline generation: A coarse-to-fine approach}.
\newblock In \emph{Proceedings of the 26th International Joint Conference on Artificial Intelligence}, IJCAI'17, page 4109–4115. AAAI Press.

\bibitem[{Team(2023)}]{geminiteam2023gemini}
Gemini Team. 2023.
\newblock \href {https://arxiv.org/abs/2312.11805} {Gemini: A family of highly capable multimodal models}.
\newblock \emph{Preprint}, arXiv:2312.11805.

\bibitem[{Tilk and Alum{\"a}e(2017)}]{tilk-alumae-2017-low}
Ottokar Tilk and Tanel Alum{\"a}e. 2017.
\newblock \href {https://doi.org/10.18653/v1/W17-4503} {Low-resource neural headline generation}.
\newblock In \emph{Proceedings of the Workshop on New Frontiers in Summarization}, pages 20--26, Copenhagen, Denmark. Association for Computational Linguistics.

\bibitem[{Touvron et~al.(2023)Touvron, Martin, Stone, Albert, Almahairi, Babaei, Bashlykov, Batra, Bhargava, Bhosale, Bikel, Blecher, Ferrer, Chen, Cucurull, Esiobu, Fernandes, Fu, Fu, Fuller, Gao, Goswami, Goyal, Hartshorn, Hosseini, Hou, Inan, Kardas, Kerkez, Khabsa, Kloumann, Korenev, Koura, Lachaux, Lavril, Lee, Liskovich, Lu, Mao, Martinet, Mihaylov, Mishra, Molybog, Nie, Poulton, Reizenstein, Rungta, Saladi, Schelten, Silva, Smith, Subramanian, Tan, Tang, Taylor, Williams, Kuan, Xu, Yan, Zarov, Zhang, Fan, Kambadur, Narang, Rodriguez, Stojnic, Edunov, and Scialom}]{touvron2023llama}
Hugo Touvron, Louis Martin, Kevin Stone, Peter Albert, Amjad Almahairi, Yasmine Babaei, Nikolay Bashlykov, Soumya Batra, Prajjwal Bhargava, Shruti Bhosale, Dan Bikel, Lukas Blecher, Cristian~Canton Ferrer, Moya Chen, Guillem Cucurull, David Esiobu, Jude Fernandes, Jeremy Fu, Wenyin Fu, Brian Fuller, Cynthia Gao, Vedanuj Goswami, Naman Goyal, Anthony Hartshorn, Saghar Hosseini, Rui Hou, Hakan Inan, Marcin Kardas, Viktor Kerkez, Madian Khabsa, Isabel Kloumann, Artem Korenev, Punit~Singh Koura, Marie-Anne Lachaux, Thibaut Lavril, Jenya Lee, Diana Liskovich, Yinghai Lu, Yuning Mao, Xavier Martinet, Todor Mihaylov, Pushkar Mishra, Igor Molybog, Yixin Nie, Andrew Poulton, Jeremy Reizenstein, Rashi Rungta, Kalyan Saladi, Alan Schelten, Ruan Silva, Eric~Michael Smith, Ranjan Subramanian, Xiaoqing~Ellen Tan, Binh Tang, Ross Taylor, Adina Williams, Jian~Xiang Kuan, Puxin Xu, Zheng Yan, Iliyan Zarov, Yuchen Zhang, Angela Fan, Melanie Kambadur, Sharan Narang, Aurelien Rodriguez, Robert Stojnic, Sergey Edunov, and Thomas
  Scialom. 2023.
\newblock \href {https://arxiv.org/abs/2307.09288} {Llama 2: Open foundation and fine-tuned chat models}.
\newblock \emph{Preprint}, arXiv:2307.09288.

\bibitem[{Verma et~al.(2023)Verma, Jangra, Verma, and Saha}]{m3ls-verma-etal-2023-large}
Yash Verma, Anubhav Jangra, Raghvendra Verma, and Sriparna Saha. 2023.
\newblock \href {https://doi.org/10.18653/v1/2023.eacl-main.263} {Large scale multi-lingual multi-modal summarization dataset}.
\newblock In \emph{Proceedings of the 17th Conference of the European Chapter of the Association for Computational Linguistics}, pages 3620--3632, Dubrovnik, Croatia. Association for Computational Linguistics.

\bibitem[{Wei et~al.(2022)Wei, Bosma, Zhao, Guu, Yu, Lester, Du, Dai, and Le}]{wei2022finetuned}
Jason Wei, Maarten Bosma, Vincent Zhao, Kelvin Guu, Adams~Wei Yu, Brian Lester, Nan Du, Andrew~M. Dai, and Quoc~V Le. 2022.
\newblock \href {https://openreview.net/forum?id=gEZrGCozdqR} {Finetuned language models are zero-shot learners}.
\newblock In \emph{International Conference on Learning Representations}.

\bibitem[{Witten et~al.(1999)Witten, Paynter, Frank, Gutwin, and Nevill-Manning}]{witten1999kea}
Ian~H. Witten, Gordon~W. Paynter, Eibe Frank, Carl Gutwin, and Craig~G. Nevill-Manning. 1999.
\newblock \href {https://arxiv.org/abs/cs/9902007} {Kea: Practical automatic keyphrase extraction}.
\newblock \emph{Preprint}, arXiv:cs/9902007.

\bibitem[{Wolf et~al.(2020)Wolf, Debut, Sanh, Chaumond, Delangue, Moi, Cistac, Rault, Louf, Funtowicz, Davison, Shleifer, von Platen, Ma, Jernite, Plu, Xu, Scao, Gugger, Drame, Lhoest, and Rush}]{wolf-etal-2020-transformers}
Thomas Wolf, Lysandre Debut, Victor Sanh, Julien Chaumond, Clement Delangue, Anthony Moi, Pierric Cistac, Tim Rault, Rémi Louf, Morgan Funtowicz, Joe Davison, Sam Shleifer, Patrick von Platen, Clara Ma, Yacine Jernite, Julien Plu, Canwen Xu, Teven~Le Scao, Sylvain Gugger, Mariama Drame, Quentin Lhoest, and Alexander~M. Rush. 2020.
\newblock \href {https://www.aclweb.org/anthology/2020.emnlp-demos.6} {Transformers: State-of-the-art natural language processing}.
\newblock In \emph{Proceedings of the 2020 Conference on Empirical Methods in Natural Language Processing: System Demonstrations}, pages 38--45, Online. Association for Computational Linguistics.

\bibitem[{Xu et~al.(2019)Xu, Wu, Madotto, and Fung}]{xu-etal-2019-clickbait}
Peng Xu, Chien-Sheng Wu, Andrea Madotto, and Pascale Fung. 2019.
\newblock \href {https://doi.org/10.18653/v1/D19-1303} {Clickbait? sensational headline generation with auto-tuned reinforcement learning}.
\newblock In \emph{Proceedings of the 2019 Conference on Empirical Methods in Natural Language Processing and the 9th International Joint Conference on Natural Language Processing (EMNLP-IJCNLP)}, pages 3065--3075, Hong Kong, China. Association for Computational Linguistics.

\bibitem[{Xue et~al.(2021)Xue, Constant, Roberts, Kale, Al-Rfou, Siddhant, Barua, and Raffel}]{xue-etal-2021-mt5}
Linting Xue, Noah Constant, Adam Roberts, Mihir Kale, Rami Al-Rfou, Aditya Siddhant, Aditya Barua, and Colin Raffel. 2021.
\newblock \href {https://doi.org/10.18653/v1/2021.naacl-main.41} {m{T}5: A massively multilingual pre-trained text-to-text transformer}.
\newblock In \emph{Proceedings of the 2021 Conference of the North American Chapter of the Association for Computational Linguistics: Human Language Technologies}, pages 483--498, Online. Association for Computational Linguistics.

\bibitem[{Yu et~al.(2022)Yu, Zhu, Li, Hu, Wang, Ji, and Jiang}]{Knowledge-Enhanced-Text-Generation}
Wenhao Yu, Chenguang Zhu, Zaitang Li, Zhiting Hu, Qingyun Wang, Heng Ji, and Meng Jiang. 2022.
\newblock \href {https://doi.org/10.1145/3512467} {A survey of knowledge-enhanced text generation}.
\newblock \emph{ACM Comput. Surv.}, 54(11s).

\bibitem[{Yuan et~al.(2020)Yuan, Wang, Meng, Thaker, Brusilovsky, He, and Trischler}]{yuan-etal-2020-one}
Xingdi Yuan, Tong Wang, Rui Meng, Khushboo Thaker, Peter Brusilovsky, Daqing He, and Adam Trischler. 2020.
\newblock \href {https://doi.org/10.18653/v1/2020.acl-main.710} {One size does not fit all: Generating and evaluating variable number of keyphrases}.
\newblock In \emph{Proceedings of the 58th Annual Meeting of the Association for Computational Linguistics}, pages 7961--7975, Online. Association for Computational Linguistics.

\bibitem[{Zhang and Yang(2023)}]{zhang-yang-2023-mediahg}
Boning Zhang and Yang Yang. 2023.
\newblock \href {https://doi.org/10.18653/v1/2023.emnlp-main.352} {{M}edia{HG}: Rethinking eye-catchy features in social media headline generation}.
\newblock In \emph{Proceedings of the 2023 Conference on Empirical Methods in Natural Language Processing}, pages 5766--5777, Singapore. Association for Computational Linguistics.

\bibitem[{Zhang et~al.(2018)Zhang, Guo, Fan, Lan, Xu, Cao, and Cheng}]{zhang2018question}
Ruqing Zhang, Jiafeng Guo, Yixing Fan, Yanyan Lan, Jun Xu, Huanhuan Cao, and Xueqi Cheng. 2018.
\newblock \href {https://doi.org/10.1145/3269206.3271711} {Question headline generation for news articles}.
\newblock In \emph{Proceedings of the 27th ACM International Conference on Information and Knowledge Management}, CIKM '18, page 617–626, New York, NY, USA. Association for Computing Machinery.

\bibitem[{Zhang* et~al.(2020)Zhang*, Kishore*, Wu*, Weinberger, and Artzi}]{bert-score}
Tianyi Zhang*, Varsha Kishore*, Felix Wu*, Kilian~Q. Weinberger, and Yoav Artzi. 2020.
\newblock \href {https://openreview.net/forum?id=SkeHuCVFDr} {Bertscore: Evaluating text generation with bert}.
\newblock In \emph{International Conference on Learning Representations}.

\bibitem[{Zhao et~al.(2024)Zhao, Zhang, Gao, Zhang, Gui, and Huang}]{zhao2024llama}
Jun Zhao, Zhihao Zhang, Luhui Gao, Qi~Zhang, Tao Gui, and Xuanjing Huang. 2024.
\newblock \href {https://arxiv.org/abs/2401.01055} {Llama beyond english: An empirical study on language capability transfer}.
\newblock \emph{Preprint}, arXiv:2401.01055.

\bibitem[{Zhao et~al.(2023)Zhao, Chen, Wang, Jiao, Long, Qin, Ding, Guo, Li, Li, and Joty}]{zhao-etal-2023-retrieving}
Ruochen Zhao, Hailin Chen, Weishi Wang, Fangkai Jiao, Do~Long, Chengwei Qin, Bosheng Ding, Xiaobao Guo, Minzhi Li, Xingxuan Li, and Shafiq Joty. 2023.
\newblock \href {https://aclanthology.org/2023.findings-emnlp.314} {Retrieving multimodal information for augmented generation: A survey}.
\newblock In \emph{Findings of the Association for Computational Linguistics: EMNLP 2023}, pages 4736--4756, Singapore. Association for Computational Linguistics.

\bibitem[{Zhou et~al.(2017)Zhou, Yang, Wei, and Zhou}]{zhou2017selective}
Qingyu Zhou, Nan Yang, Furu Wei, and Ming Zhou. 2017.
\newblock \href {https://doi.org/10.18653/v1/P17-1101} {Selective encoding for abstractive sentence summarization}.
\newblock In \emph{Proceedings of the 55th Annual Meeting of the Association for Computational Linguistics (Volume 1: Long Papers)}, pages 1095--1104, Vancouver, Canada. Association for Computational Linguistics.

\bibitem[{Zhu et~al.(2018)Zhu, Li, Liu, Zhou, Zhang, and Zong}]{zhu-etal-2018-msmo}
Junnan Zhu, Haoran Li, Tianshang Liu, Yu~Zhou, Jiajun Zhang, and Chengqing Zong. 2018.
\newblock \href {https://doi.org/10.18653/v1/D18-1448} {{MSMO}: Multimodal summarization with multimodal output}.
\newblock In \emph{Proceedings of the 2018 Conference on Empirical Methods in Natural Language Processing}, pages 4154--4164, Brussels, Belgium. Association for Computational Linguistics.

\bibitem[{Zhu et~al.(2019)Zhu, Wang, Wang, Zhou, Zhang, Wang, and Zong}]{zhu-etal-2019-ncls}
Junnan Zhu, Qian Wang, Yining Wang, Yu~Zhou, Jiajun Zhang, Shaonan Wang, and Chengqing Zong. 2019.
\newblock \href {https://doi.org/10.18653/v1/D19-1302} {{NCLS}: Neural cross-lingual summarization}.
\newblock In \emph{Proceedings of the 2019 Conference on Empirical Methods in Natural Language Processing and the 9th International Joint Conference on Natural Language Processing (EMNLP-IJCNLP)}, pages 3054--3064, Hong Kong, China. Association for Computational Linguistics.

\end{thebibliography}
\clearpage
\appendix
\twocolumn[{%
 \centering
 \Large\bf Supplementary Material: Appendices \\ [20pt]
}]


\section{Data Processing \& Statistics}
\label{sec:dataset-processing}

\subsection{Data Processing}
The data processing involved the application of regular expressions \texttt{Regex}\footnote{\url{https://docs.python.org/3/library/re.html}} to eliminate URLs, HTML tags, and emojis from the headline, article, and image captions. Furthermore, \texttt{Phonenumbers} package\footnote{\url{https://pypi.org/project/phonenumbers/}} was employed to specifically filter out phone numbers from the textual content.

\subsection{Data Statistics}
\label{subsec:data-statistics}
Following data processing, the dataset comprises 415,117 news samples, represented as tuples encompassing headline, article , images, image captions, and tag words. The task of headline generation is herein considered as a subproblem integral to extreme summarization. Given the dual nature of summarization extractive and abstractive, our objective is to craft abstractive headlines. In the pursuit of measuring abstractiveness, \citet{narayan2018don} proposed evaluating the percentage of n-grams in the summary that do not find occurrence in the input article. Simultaneously, \citet{bommasani2020intrinsic} introduced compression as a metric for quantifying conciseness, expressed mathematically as:

\begin{equation}
\textbf{Compression}(H, A) = 1 - \frac{|H|}{|A|}
\end{equation}

In this context, $|H|$ and $|A|$ represent the lengths of the headline and article, measured in words. The ideal headline is aimed to be informative, concise, and have a high ratio of novel n-grams.

The process of generating tag words is in line with Keyphrase generation. Extracting tag words seeks to capture existing words in the input document. However, it's crucial to note that extractive methods can't predict absent tag words, an important aspect of comprehensive tag word generation. In contrast, tag word generation aims to produce both present and absent tag words \cite{meng-etal-2017-deep}. The percentage of present tag words highlights the challenge of predicting tag words not present in the document.

In Table~\ref{table:datas-stats-1}, we present a comprehensive set of quantitative statistics that includes both the previously mentioned metrics and additional ones. The average word count in articles is 902, surpassing the Transformer Encoder's contextual window capacity. Using the mt5 multilingual sentencepiece tokenizer with a vocabulary size of 250,112, the mean token count per article to 1631. This is noteworthy considering the mt5 encoder's context window is limited to 512 tokens, leading to truncation beyond this limit. This emphasizes the need for methodologies capable of capturing essential information from documents exceeding the encoder context limit.

Further analysis shows that, on average, articles require compression of 98.88\% for effective headline representation, highlighting the significant challenge in abstractive headline generation. Additionally, we observe that the percentage of tag words in the document consistently falls below 50\%, emphasizing the importance of generative approaches for keywords to overcome limitations posed by extraction methods. These findings enhance our understanding of the intricate dynamics of document structure and the challenges associated with information extraction and summarization.

\section{Handling Multiple Images and Captions}
\label{sec:greddy-algo}

The primary distinction between our modules lies in the modalities of the auxiliary data they utilize. As discussed in Section~\ref{sec:retriever}, our goal is to retrieve salient and pertinent information from news articles. To achieve this, we developed two separate modules for retrieving the most relevant sentences, each leveraging auxiliary information from different modalities. Specifically, \texttt{\textbf{ImgRet}} uses images, and \texttt{\textbf{CapRet}} uses image captions to retrieve the most relevant sentences from a document, respectively.

We observed that a single document often contains multiple images and captions without a proper one-to-one mapping between them. Consequently, we treat each image and caption as distinct entities and propose a greedy algorithm for aggregating multiple retrievals. The following discussion predominantly focuses on the operations of the \texttt{\textbf{ImgRet}} module, with selective explication of the points where \texttt{\textbf{CapRet}} differs.

The \textbf{ImgRet} module processes input data consisting of the \underline{$Article$}, \underline{$Language$}, \underline{$Images$}, and a parameter denoted as \underline{$K$} (indicating the top-k sentences for retrieval). Subsequently, the \textbf{SentenceSegmenter} is utilized to deconstruct the \underline{$Article$} into individual sentences, employing both the \underline{$Article$} and \underline{$Language$} as inputs. Subsequent steps involve iteratively comparing each sentence with the images, accumulating similarity scores for each sentence. The sentences are then sorted based on these scores, and the top \underline{$K$} sentences, representing the most similar ones, are selected. The final sorting is performed based on their original order in the \underline{$Article$}, yielding the most relevant sentences in their original sequence.

It is noteworthy that \textbf{CapRet} diverges from this procedure solely in the calculation of embeddings for image captions, as opposed to the image embeddings computed in the \textbf{ImgRet} module.

\section{Multilingual Tools}
\label{sec:multilingual-tools}
The goal of \textbf{\texttt{Multilingual Sentence Tokenizer}} is to split a provided document into separate sentences. The language scope and the open-source resources employed for this tool are detailed in Table \ref{table:multilingual-sentence-tokenizer}.

\textbf{\texttt{Multilingual Stemmer}} is used for normalizing tag words. The languages supported and their respective implementation sources are outlined in Table \ref{table:multilingual-stemmer}.
\begin{table}[ht]
\centering
\begin{tabular}{l|l||ll}
\Xhline{4\arrayrulewidth}
\multicolumn{1}{c|}{\textbf{Language}} & \multicolumn{1}{c||}{\textbf{Source}} & \multicolumn{1}{c|}{\textbf{Language}} & \multicolumn{1}{c}{\textbf{Source}} \\ \Xhline{4\arrayrulewidth}
English                                & \multirow{3}{*}{\href{https://www.nltk.org/}{NLTK}}                & \multicolumn{1}{l|}{Slovak}            & \href{https://github.com/nipunsadvilkar/pySBD}{PYSBD}                               \\ \cline{1-1} \cline{3-4} 
Portuguese                             &                                      & \multicolumn{1}{l|}{Indonesian}        & \multirow{9}{*}{\href{https://spacy.io/}{Spacy}}              \\ \cline{1-1} \cline{3-3}
Turkish                                &                                      & \multicolumn{1}{l|}{Nepali}            &                                     \\ \cline{1-3}
Arabic                                 & \multirow{18}{*}{\href{https://github.com/nipunsadvilkar/pySBD}{PYSBD}}              & \multicolumn{1}{l|}{Ukrainian}         &                                     \\ \cline{1-1} \cline{3-3}
French                                 &                                      & \multicolumn{1}{l|}{Chinese}           &                                     \\ \cline{1-1} \cline{3-3}
Spanish                                &                                      & \multicolumn{1}{l|}{Yoruba}            &                                     \\ \cline{1-1} \cline{3-3}
Persian                                &                                      & \multicolumn{1}{l|}{Vietnamese}        &                                     \\ \cline{1-1} \cline{3-3}
Russian                                &                                      & \multicolumn{1}{l|}{Thai}              &                                     \\ \cline{1-1} \cline{3-3}
Urdu                                   &                                      & \multicolumn{1}{l|}{Slovenian}         &                                     \\ \cline{1-1} \cline{3-3}
Amharic                                &                                      & \multicolumn{1}{l|}{Sinhala}           &                                     \\ \cline{1-1} \cline{3-4} 
Armenian                               &                                      & \multicolumn{1}{l|}{Bengali}           & \href{https://pypi.org/project/bltk/}{BLTK}                                \\ \cline{1-1} \cline{3-4} 
Bulgarian                              &                                      & \multicolumn{1}{l|}{Gujarati}          & \multirow{9}{*}{\href{https://github.com/anoopkunchukuttan/indic_nlp_library}{IndicNLP}}              \\ \cline{1-1} \cline{3-3}
Polish                                 &                                      & \multicolumn{1}{l|}{Hindi}             &                                     \\ \cline{1-1} \cline{3-3}
Dutch                                  &                                      & \multicolumn{1}{l|}{Marathi}           &                                     \\ \cline{1-1} \cline{3-3}
Danish                                 &                                      & \multicolumn{1}{l|}{Punjabi}           &                                     \\ \cline{1-1} \cline{3-3}
Burmese                                &                                      & \multicolumn{1}{l|}{Tamil}             &                                     \\ \cline{1-1} \cline{3-3}
Greek                                  &                                      & \multicolumn{1}{l|}{Telugu}            &                                     \\ \cline{1-1} \cline{3-3}
Italian                                &                                      & \multicolumn{1}{l|}{Oriya}             &                                     \\ \cline{1-1} \cline{3-3}
Japanese                               &                                      & \multicolumn{1}{l|}{kannada}           &                                     \\ \cline{1-1} \cline{3-3}
German                                 &                                      & \multicolumn{1}{l|}{Malayalam}         &                                     \\ \cline{1-1} \cline{3-4} 
Kazakh                                 &                                      &                                        &                                     \\ \cline{1-2}
\end{tabular}
\caption{Multilingual Sentence Tokenizer supports \num{41} different languages.}
\label{table:multilingual-sentence-tokenizer}
\end{table}

\begin{table}[ht]
\centering
\begin{tabular}{l|l}
\Xhline{4\arrayrulewidth}
\multicolumn{1}{c|}{\textbf{Language}} & \multicolumn{1}{c}{\textbf{Source}}                         \\ \Xhline{4\arrayrulewidth}
English                                & \multirow{8}{*}{\href{https://github.com/snowballstem/snowball}{Snowball}}   \\ \cline{1-1}
Portuguese                             &                                                             \\ \cline{1-1}
Spanish                                &                                                             \\ \cline{1-1}
Russian                                &                                                             \\ \cline{1-1}
French                                 &                                                             \\ \cline{1-1}
Arabic                                 &                                                             \\ \cline{1-1}
Tamil                                  &                                                             \\ \cline{1-1}
Indonesian                             &                                                             \\ \hline
Turkish                                & \href{https://github.com/hanefi/turkish-stemmer-python}{turkish-stemmer-python}            \\ \hline
Ukrainian                              & \href{https://github.com/Amice13/ukr\_stemmer}{ukr\_stemmer}                     \\ \hline
Bengali                                & \href{https://github.com/MIProtick/Bangla-stemmer}{Bangla-stemmer}                 \\ \hline
Persian                                & \href{https://github.com/htaghizadeh/PersianStemmer-Python}{PersianStemmer-Python}        \\ \hline
Nepali                                 & \href{https://github.com/oya163/nepali-stemmer}{nepali-stemmer}                   \\ \hline
Urdu                                   & \href{https://github.com/burhanharoon/Urdu-Stemmer}{Urdu-Stemmer}              \\ \hline
Gujarati                               & \href{https://github.com/Rutvik-Trivedi/Gujarati-NLP-Toolkit}{Gujarati-NLP-Toolkit}     \\ \hline
Hindi                                  & \href{https://research.variancia.com/hindi\_stemmer/}{hindi\_stemmer}             \\ \hline
Marathi                                & \href{https://github.com/sahil-zngr/IR/tree/main/marathi\_stemmer}{marathi\_stemmer} \\ \hline
Panjabi                                & \href{https://github.com/SimranKaur-23/Stemmer-in-punjabi}{Stemmer-in-punjabi}        \\ \Xhline{4\arrayrulewidth}
\end{tabular}
\caption{Multilingual Stemmer Covers 18 Language.}
\label{table:multilingual-stemmer}
\end{table}

\section{Baselines}
\label{sec:baselines}

\subsection{Headline Generation Baselines}

\paragraph{LEAD-1} In the field of generating news headlines, we often use a common benchmark called \textbf{LEAD-1} to set the minimum standard for the task \cite{kumar2022indicnlg, narayan2018don}. To calculate LEAD-1 scores, we take the first sentence of the article as the system-generated headline and compare it with the original headline.

\paragraph{EXT-ORACLE} On the flip side, there's another approach called \textbf{EXT-ORACLE}, which represents the best possible outcome when generating headlines using an extractive method \cite{kumar2022indicnlg, narayan2018don}. In this case, we align a sentence from the input article with the reference headline, using the ROUGE-2 metric to assess how well they match up.

\subsection{Tags Generation Baselines}

\paragraph{TF-IDF} approach \cite{tfidf} entails the ranking of extracted noun phrase candidates based on their term frequency and inverse document frequency within the provided documents.

\paragraph{TextRank} In the case of \textbf{TextRank} \cite{mihalcea-tarau-2004-textrank}, the methodology involves the conceptualization of words as web pages, followed by the application of the PageRank algorithm to identify keyphrases.

\paragraph{KEA} On the other hand, \textbf{KEA} \cite{witten1999kea} employs lexical methods to identify candidate keyphrases, computes feature values for each candidate, and utilizes a machine learning algorithm to predict the suitability of candidates as keyphrases.

\section{Training \& Hyperparameters}
\label{sec:hyperparameter}

Given the computational constraints and resource limitations, we chose to work with the base versions of the models for our experiments. The fine-tuning process for the mT5-base and mT0-base models was carried out on a single RTX 3090 GPU, while the Flan-T5-large model was fine-tuned using an RTX A6000 GPU. We configured the encoder input token sequence length to 512 and the decoder output token sequence length to 64.

The fine-tuning phase for the mT5-base, mT0-base, and Flan-T5-large models lasted for five epochs, employing a batch size of 8. We utilized the AdamW optimizer \cite{adam-w-loshchilov2019decoupled} throughout the training process to optimize the models. This setup was chosen to balance the trade-off between computational efficiency and model performance, ensuring effective utilization of available hardware resources.
\section{Performance of LLMs}
\label{sec:LLM-Performance}

Gemini \cite{geminiteam2023gemini} and Mixtral \cite{jiang2023mistral} are two of large language models used for various tasks. Google provides access to Gemini via an API\footnote{\href{https://ai.google.dev/}{Gemini Pro API}} for free, while the Mistral AI Team has made Mixtral's weights\footnote{\href{https://huggingface.co/mistralai/Mixtral-8x7B-v0.1}{Mixtral}} publicly available. We utilized Gemini-Pro and Mixtral models to assess their performance in a multilingual task called XL-HeadTags. This evaluation was conducted under zero-shot prompting conditions, where only input and output formats were provided as instructions. We examined 50 examples from each language.

\begin{tcolorbox}[breakable,title={\small Prompt for LLMs}]
\footnotesize
\textbf{\underline{Instruction}}
\vspace{1mm}

Generate Headline and some defined or undefined number for tag words from a given news article. When ask to generate defined number of tag words generate exact defined number of tag words seperated by commas (,). 
\vspace{2mm}

A single tag word can be multiple word or single word. When asked to generate undefined number of tag words, you should generate the number of tag word you think appropriate. 
\vspace{2mm}

While generating headline and tag words generate them in the language the article is given in. 
\vspace{2mm}

\textbf{\underline{Exemplars}}
\vspace{1mm}

For example, you will be given task in these input format "Generate Headline and Three Tag Words": "News article". \\
Output should be "Headline is: Generated Headline. Tag Words are: Tag1, Tag2, Tag3". 
\vspace{2mm}

Another example "Generate Headline and Tag Words": "News article". Your output should be "Headline is: Generated Headline. Tag Words are: Tag1, Tag2, . . . , TagN".\\

\textbf{\underline{Input:}} \{...\}
\end{tcolorbox}

However, our analysis revealed that their performance is subpar compared to a supervised approach. Several factors may have contributed to this. Firstly, both models heavily favor English. Even when presented with non-English articles and instructed to generate in the respective languages, they often produce results in English due to the dominance of English in their pre-training corpus.

Additionally, Mistral occasionally generates output in a romanized version of the target language, indicating a lack of effort in adapting to non-English contexts. We also encountered issues where the models inconsistently generated headlines and tags in different languages or failed to follow instructions, resulting in excessive tag words or overly lengthy headlines.

\begin{figure*}[t]
    \centering
    \includegraphics[scale = 0.8]{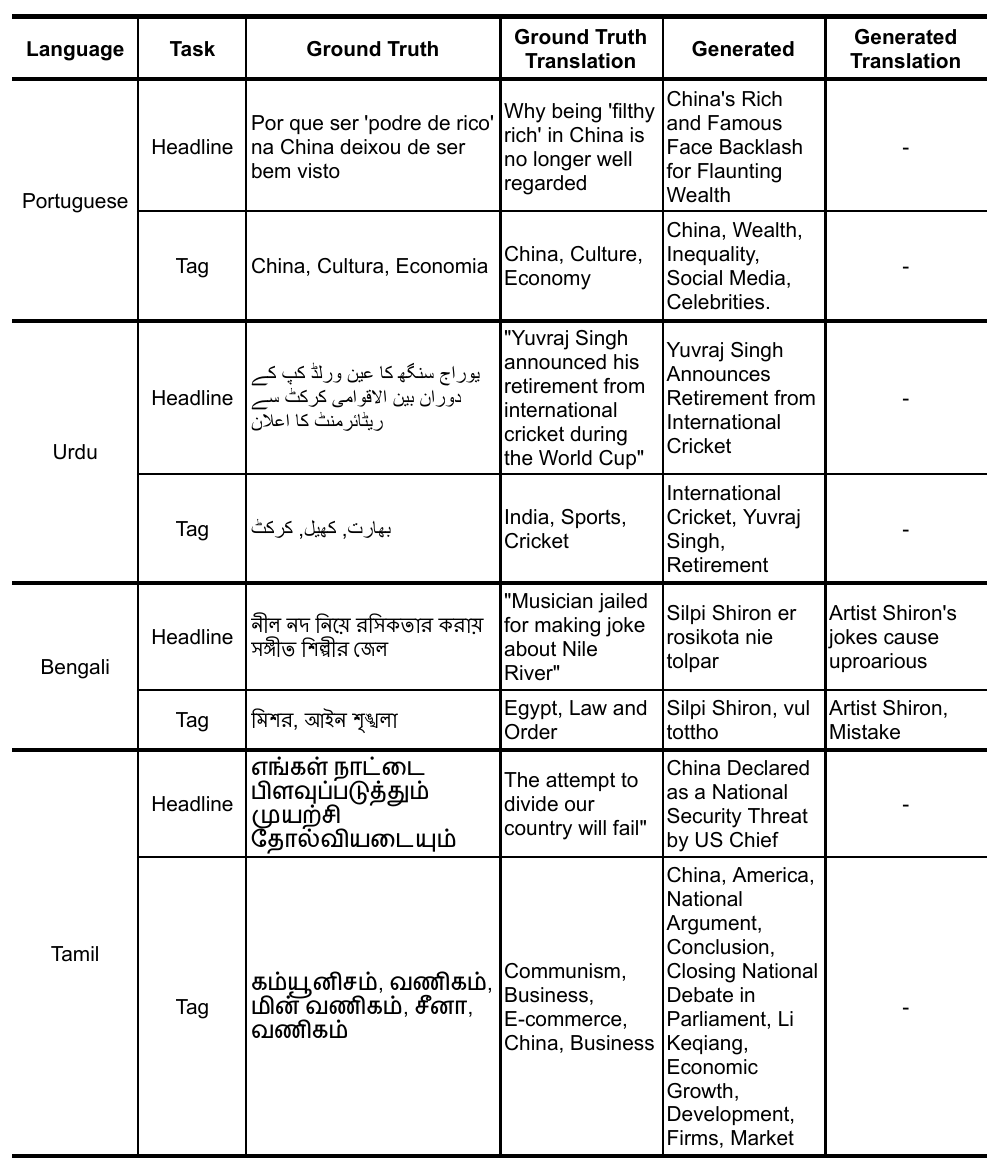}
    \caption{Large Language Models (LLMs) generated samples.}
    \label{fig:llm-generation}
\end{figure*}
\section{Detailed Related Work}
\subsection{Headline Generation}

Headline generation, a niche within abstractive summarization \cite{chali-etal-2017-towards, nayeem-chali-2017-extract, FUAD2019216, 10.1145/3132847.3133106, nayeem-etal-2019-diverse}, particularly in multilingual contexts \cite{chowdhury-etal-2021-unsupervised}, has undergone notable advancements. In English, \citet{rush2015neural} introduced an attention-based neural network model that employs a recurrent neural network (RNN) and attention mechanism for concise summarization. \citet{takase2016neural} developed an Abstract Meaning Representation (AMR) encoder within an encoder-decoder framework for headline creation, while \citet{zhang2018question} proposed a dual-attention sequence-to-sequence model tailored for question-driven headline generation. In contexts with limited resources, \citet{tilk-alumae-2017-low} demonstrated the efficacy of pre-training neural models to enhance headline generation capabilities.

\citet{zhou2017selective} approach to headline generation encompasses a tri-phase process involving sentence encoding, sentence selection via a gate network, and headline decoding. Similarly, \citet{tan2017neural} introduced a method that prioritizes important sentences for context-based headline generation. On the multilingual front, \citet{kumar2022indicnlg} launched \texttt{IndicNLG}, a dataset aimed at headline generation, though it primarily focuses on headline-article pairs without incorporating auxiliary information.

Emerging methodologies in headline generation also explore specialized applications, such as generating headlines for community question answering \cite{higurashi-etal-2018-extractive} and creating multiple headlines for varied contexts \cite{iwama-kano-2019-multiple}, broadening the scope of research in this field.

\subsection{Keyphrases Generation}
While "Tag Words" and "Keyphrases" both serve to enhance information retrieval, their application and research emphasis differ significantly. Tag word generation remains underexplored compared to the more developed field of keyphrase generation. For example, \citet{meng-etal-2017-deep} unveiled the CopyRNN model, which leverages an attentional encoder-decoder architecture \cite{bahdanau2014neural} integrated with a copying mechanism \cite{gu-etal-2016-incorporating} to predict keyphrases. This method employs over-generation with an extensive beam search to then select the top N predictions, such as the top five or ten keyphrases.

\citet{yuan-etal-2020-one} took an alternative approach by designing a Keyphrase Generation model that not only predicts multiple keyphrases but also determines the appropriate number of keyphrases for each document. This innovative training scheme enables the model to adaptively generate a variable number of keyphrases tailored to the specifics of each document. Further advancing the field, \citet{chan-etal-2019-neural} introduced the application of Reinforcement Learning to optimize the keyphrase generation process, showcasing the evolving techniques aimed at improving the precision and adaptability of keyphrase generation.

\subsection{Multitask Learning and Instruction Tuning}

In the realm of NLP, multitask learning (MTL) is increasingly recognized for its potential to enhance model performance on related tasks by exploiting their shared features and distinctions. For example, \citet{10.5555/3298483.3298678} developed a model that simultaneously trains on summary generation and text classification, achieving notable improvements in text summarization. \citet{dong-etal-2015-multi} pioneered a multitask learning approach using a sequence-to-sequence (Seq2Seq) framework for translating a single source language into multiple target languages. \citet{zhu-etal-2019-ncls} explored the synergies between monolingual summarization, machine translation, and cross-lingual summarization through joint training. Similarly, \citet{guo-etal-2018-soft} proposed a model that concurrently learns abstractive summarization and question generation. Extending the scope of multitask learning, \citet{sanh2022multitask} demonstrated the effectiveness of prompted multi-task fine-tuning on a pre-trained T5 model \cite{raffel2020t5} for zero-shot task generalization.

Furthermore, language models can undergo fine-tuning on supervised datasets comprising natural language prompts paired with their respective target outputs. This method, termed \textbf{``instruction tuning,''} significantly refines the models' proficiency in adhering to instructions. Employing task-specific prefixes, this technique directs the model to generate outputs in a predetermined format, showcasing the versatility and precision attainable through instruction-based training.

\subsection{Retrieval-Augmented Generation}

Retrieval-Augmented Generation (RAG) represents a pivotal advancement in Natural Language Generation (NLG), addressing the issue of neural models' limited contextual understanding. Traditional neural models often falter when the input lacks comprehensive information for generating accurate outputs, particularly in complex real-world applications \cite{Knowledge-Enhanced-Text-Generation}. To bridge this gap, KNNLM \cite{khandelwal2020generalization} introduced a technique for augmenting language models with examples retrieved from a training text dataset, enhancing contextual relevance. Building on this, RETRO \cite{borgeaud2021improving} leveraged a vastly expanded text corpus, enabling models with a smaller footprint to achieve performance on par with GPT-3 \cite{brown2020language}.

Models such as REALM \cite{pmlr-v119-guu20a} and RAG \cite{lewis2020retrieval} incorporate Wikipedia passages as external knowledge bases, significantly boosting their efficacy in tasks like Question Answering. REALM focuses on encoding information through masked language modeling, whereas RAG employs an encoder-decoder structure for generative language tasks.

Expanding on these concepts, MuRAG \cite{chen-etal-2022-murag} stands out by integrating multimodal knowledge sources, encompassing both visual and textual data. This innovation extends the capabilities of knowledge-enhanced text generation, catering to the nuanced demands of intricate information landscapes.

\begin{table*}[t!]
\centering
\renewcommand{\arraystretch}{1} 
\resizebox{16cm}{!} 
{ 
\begin{tabular}{ccccccccc}
\Xhline{4\arrayrulewidth}
\multicolumn{2}{c}{}                                                                                     & \textbf{R1}                   & \textbf{R2}                   & \textbf{RL}                   & \textbf{BLEU}                 & \textbf{METEOR}               & \textbf{LR}                  & \textbf{BERT Score}                \\ \Xhline{4\arrayrulewidth}
\multicolumn{2}{c|}{\textbf{Models}}                                                                     & \multicolumn{7}{c}{\textbf{Baselines}}                                                                                                                                                                                       \\ \hline
\rowcolor[HTML]{DCDCDC} 
\multicolumn{2}{c|}{\cellcolor[HTML]{DCDCDC}mT5}                                                         & 37.86                         & 17.20                         & 33.53                         & 12.95                         & 25.55                         & 0.84                         & 75.79                         \\
\multicolumn{2}{c|}{mT0}                                                                                 & 38.33                         & 17.66                         & 33.90                         & 14.64                         & 26.44                         & 0.94                         & 75.83                         \\
\rowcolor[HTML]{DCDCDC} 
\multicolumn{2}{c|}{\cellcolor[HTML]{DCDCDC}FLAN-T5}                                                     & 31.46                         & 12.73                         & 28.15                         & 8.75                          & 24.61                         & 0.71                         & 70.87                         \\
\end{tabular}
}
\caption{Headline Generation Evaluation. \textbf{Selected Content} (Only Important Sentences). The best results compared to their respective baseline models are marked in \textbf{bold}, and $\Delta$ gains are shown in round brackets and highlighted with \textcolor{darkgreen}{green} and \textcolor{customRed}{red} colors.
\label{table:headline-evaluation}}
\end{table*}
\begin{table*}[ht]
\centering
\renewcommand{\arraystretch}{1} 
\resizebox{11cm}{!} 
{ 
\begin{tabular}{cccccc}
\Xhline{4\arrayrulewidth}
\multicolumn{2}{c}{}                                                                                    & \ft            & \ff            & \fm            & \fo            \\ \Xhline{4\arrayrulewidth}
\multicolumn{2}{c|}{\textbf{Models}}                                                                     & \multicolumn{4}{c}{\textbf{Baselines}}                                                                                                                                                                                    \\ \hline
\rowcolor[HTML]{E8E8E8} 
\multicolumn{2}{c|}{\cellcolor[HTML]{E8E8E8}mT5}                                                        & 45.01                         & 39.82                         & 44.67                         & 46.79                         \\
\multicolumn{2}{c|}{mT0}                                                                                & 51.58                         & 44.94                         & 52.50                         & 54.39                         \\
\rowcolor[HTML]{E8E8E8} 
\multicolumn{2}{c|}{\cellcolor[HTML]{E8E8E8}Flan-T5}                                                                                & 30.76                         & 26.30                         & 31.86                         & 33.40           \\
\multicolumn{2}{c|}{Gemini-Pro}                                                                          & 5.90                                                    & 7.75                                                    & 6.57                                                    & 6.65                                                    \\
\end{tabular}
}
\caption{Tag Words Evaluation. \textbf{Selected Content} (Important Sentence + Article). The best results compared to their respective baseline models are marked in \textbf{bold}, and $\Delta$ gains are shown in round brackets and highlighted with \textcolor{darkgreen}{green} and \textcolor{customRed}{red} colors.
\label{table:tags-evaluation-ia}}
\end{table*}

\begin{table*}[ht]
\resizebox{16.2cm}{!} 
{ 
\centering
\renewcommand{\arraystretch}{1} 
\begin{tabular}{cccccccccccccc}
\hline
                                            &                                                         & \multicolumn{12}{c}{\textbf{Text} (\textit{Caption})}                                \\ \cline{3-14} 

                                            &                                                         & \multicolumn{4}{c|}{\textbf{-w/C (K=5)}}                                                                                                           & \multicolumn{4}{c|}{\textbf{-w/C (K=10)}}                                                                                                          & \multicolumn{4}{c}{\textbf{-w/C (K=15)}}                                                                                      \\ \hline
\textbf{Resources}                               & \textbf{Languages}                                      & \textbf{R2}                   & \textbf{RL}                   & \textbf{BLEU}                 & \multicolumn{1}{c|}{\textbf{METEOR}}                 & \textbf{R2}                   & \textbf{RL}                   & \textbf{BLEU}                 & \multicolumn{1}{c|}{\textbf{METEOR}}                 & \textbf{R2}                   & \textbf{RL}                   & \textbf{BLEU}                 & \textbf{METEOR}                 \\ \hline
\multicolumn{1}{c|}{}                       & \multicolumn{1}{c|}{\cellcolor[HTML]{E8E8E8}English}    & \cellcolor[HTML]{E8E8E8}16.35 & \cellcolor[HTML]{E8E8E8}36.36 & \cellcolor[HTML]{E8E8E8}14    & \multicolumn{1}{c|}{\cellcolor[HTML]{E8E8E8}30.3}  & \cellcolor[HTML]{E8E8E8}16.91 & \cellcolor[HTML]{E8E8E8}37.06 & \cellcolor[HTML]{E8E8E8}14.43 & \multicolumn{1}{c|}{\cellcolor[HTML]{E8E8E8}31.05} & \cellcolor[HTML]{E8E8E8}17.1  & \cellcolor[HTML]{E8E8E8}37.05 & \cellcolor[HTML]{E8E8E8}13.78 & \cellcolor[HTML]{E8E8E8}31.16 \\
\hline

\end{tabular}
}
\caption{Headline Generation Evaluation : \textbf{mT5}. \textbf{Selected Content} (Only Important Sentences)}
\label{table:language-headline-mt5}
\end{table*}
\begin{table*}[ht]
\resizebox{16.2cm}{!} 
{ 
\centering
\renewcommand{\arraystretch}{1} 
\begin{tabular}{cccccccccccccc}
\hline
                                            &                                                         & \multicolumn{12}{c}{\textbf{Text} (\textit{Caption})}                    \\ \cline{3-14} 

                                            &                                                         & \multicolumn{4}{c|}{\textbf{-w/C (K=5)}}                                                                                                           & \multicolumn{4}{c|}{\textbf{-w/C (K=10)}}                                                                                                          & \multicolumn{4}{c}{\textbf{-w/C (K=15)}}                                                                                      \\ \hline
\textbf{Resources}                           & \textbf{Languages}                                      & \textbf{R2}                   & \textbf{RL}                   & \textbf{BLEU}                 & \multicolumn{1}{c|}{\textbf{METEOR}}                 & \textbf{R2}                   & \textbf{RL}                   & \textbf{BLEU}                 & \multicolumn{1}{c|}{\textbf{METEOR}}                 & \textbf{R2}                   & \textbf{RL}                   & \textbf{BLEU}                 & \textbf{METEOR}                 \\ \hline
\multicolumn{1}{c|}{}                       & \multicolumn{1}{c|}{\cellcolor[HTML]{E8E8E8}English}    & \cellcolor[HTML]{E8E8E8}16.78 & \cellcolor[HTML]{E8E8E8}36.77 & \cellcolor[HTML]{E8E8E8}14.33 & \multicolumn{1}{c|}{\cellcolor[HTML]{E8E8E8}30.9}  & \cellcolor[HTML]{E8E8E8}17.55 & \cellcolor[HTML]{E8E8E8}37.56 & \cellcolor[HTML]{E8E8E8}15.03 & \multicolumn{1}{c|}{\cellcolor[HTML]{E8E8E8}31.82} & \cellcolor[HTML]{E8E8E8}17.11 & \cellcolor[HTML]{E8E8E8}37.05 & \cellcolor[HTML]{E8E8E8}13.78 & \cellcolor[HTML]{E8E8E8}31.16 \\
\hline
\end{tabular}
}
\caption{Headline Generation Evaluation : \textbf{mT0}. \textbf{Selected Content} (Only Important Sentences).}
\label{table:language-headline-mt0}
\end{table*}
\begin{table*}[ht]
\resizebox{16.2cm}{!} 
{ 
\centering
\renewcommand{\arraystretch}{1} 
\begin{tabular}{cccccccccccccc}
\hline
                                            &                                                         & \multicolumn{12}{c}{\textbf{Text} (\textit{Caption})}                    \\ \cline{3-14} 
                                            &                                                         & \multicolumn{4}{c|}{\textbf{-w/C (K=5)}}                                                                                                           & \multicolumn{4}{c|}{\textbf{-w/C (K=10)}}                                                                                                          & \multicolumn{4}{c}{\textbf{-w/C (K=15)}}                                                                                      \\ \hline
\textbf{Resources}                               & \textbf{Languages}                                      & \textbf{R2}                   & \textbf{RL}                   & \textbf{BLEU}                 & \multicolumn{1}{c|}{\textbf{METEOR}}                 & \textbf{R2}                   & \textbf{RL}                   & \textbf{BLEU}                 & \multicolumn{1}{c|}{\textbf{METEOR}}                 & \textbf{R2}                   & \textbf{RL}                   & \textbf{BLEU}                 & \textbf{METEOR}                 \\ \hline
\multicolumn{1}{c|}{}                       & \multicolumn{1}{c|}{\cellcolor[HTML]{E8E8E8}English}    & \cellcolor[HTML]{E8E8E8}20.34 & \cellcolor[HTML]{E8E8E8}40.42 & \cellcolor[HTML]{E8E8E8}17.08 & \multicolumn{1}{c|}{\cellcolor[HTML]{E8E8E8}35.45} & \cellcolor[HTML]{E8E8E8}21.31 & \cellcolor[HTML]{E8E8E8}41.5  & \cellcolor[HTML]{E8E8E8}18.08 & \multicolumn{1}{c|}{\cellcolor[HTML]{E8E8E8}36.56} & \cellcolor[HTML]{E8E8E8}21.57 & \cellcolor[HTML]{E8E8E8}41.73 & \cellcolor[HTML]{E8E8E8}18.34 & \cellcolor[HTML]{E8E8E8}36.94 \\
\hline
\end{tabular}
}
\caption{Headline Generation Evaluation : \textbf{Flan-T5}. \textbf{Selected Content} (Only Important Sentences)}
\label{table:language-headline-flant5}
\end{table*}

\begin{table*}[ht]
\resizebox{16.2cm}{!} 
{ 
\centering
\renewcommand{\arraystretch}{1} 
\begin{tabular}{cccccccccccccc}
\hline
                                            &                                                         & \multicolumn{12}{c}{\textbf{Text} (\textit{Caption})}                                                                                                                                                                                                                                                                                                                                                                                                      \\ \cline{3-14} 

                                            &                                                         & \multicolumn{4}{c|}{\textbf{-w/C (K=5)}}                                                                                                           & \multicolumn{4}{c|}{\textbf{-w/C (K=10)}}                                                                                                          & \multicolumn{4}{c}{\textbf{-w/C (K=15)}}                                                                                      \\ \hline
\textbf{Resources}                               & \textbf{Languages}                                      & \textbf{\ft}                   & \textbf{\ff}                   & \textbf{\fm}                   & \multicolumn{1}{c|}{\textbf{\fo}}                   & \textbf{\ft}                   & \textbf{\ff}                   & \textbf{\fm}                   & \multicolumn{1}{c|}{\textbf{\fo}}                   & \textbf{\ft}                   & \textbf{\ff}                   & \textbf{\fm}                   & \textbf{\fo}                   \\
\hline
\end{tabular}
}
\caption{Tags Words Evaluation : \textbf{mT5}. \textbf{Selected Content} (Only Important Sentences).}
\label{table:language-tags-mt5}
\end{table*}
\begin{table*}[ht]
\resizebox{16.2cm}{!} 
{ 
\centering
\renewcommand{\arraystretch}{1} 
\begin{tabular}{cccccccccccccc}
\hline
                                            &                                                         & \multicolumn{12}{c}{\textbf{Text} (\textit{Caption})}                                                                           \\ \cline{3-14} 

                                            &                                                         & \multicolumn{4}{c|}{\textbf{-w/C (K=5)}}                                                                                                           & \multicolumn{4}{c|}{\textbf{-w/C (K=10)}}                                                                                                          & \multicolumn{4}{c}{\textbf{-w/C (K=15)}}                                                                                      \\ \hline
\textbf{Resources}                               & \textbf{Languages}                                      & \textbf{\ft}                   & \textbf{\ff}                   & \textbf{\fm}                   & \multicolumn{1}{c|}{\textbf{\fo}}                   & \textbf{\ft}                   & \textbf{\ff}                   & \textbf{\fm}                   & \multicolumn{1}{c|}{\textbf{\fo}}                   & \textbf{\ft}                   & \textbf{\ff}                   & \textbf{\fm}                   & \textbf{\fo}                   \\ \hline
\multicolumn{1}{c|}{}                       & \multicolumn{1}{c|}{\cellcolor[HTML]{E8E8E8}English}    & \cellcolor[HTML]{E8E8E8}51.85 & \cellcolor[HTML]{E8E8E8}44.4  & \cellcolor[HTML]{E8E8E8}54.8  & \multicolumn{1}{c|}{\cellcolor[HTML]{E8E8E8}57.81} & \cellcolor[HTML]{E8E8E8}53.31 & \cellcolor[HTML]{E8E8E8}45.6  & \cellcolor[HTML]{E8E8E8}55.93 & \multicolumn{1}{c|}{\cellcolor[HTML]{E8E8E8}59.05} & \cellcolor[HTML]{E8E8E8}47.46 & \cellcolor[HTML]{E8E8E8}40.78 & \cellcolor[HTML]{E8E8E8}48.24 & \cellcolor[HTML]{E8E8E8}51.25 \\
\hline
\end{tabular}
}
\caption{Tags Words Evaluation : \textbf{mT0}. \textbf{Selected Content} (Only Important Sentences).}
\label{table:language-tags-mt0}
\end{table*}
\begin{table*}[ht]
\resizebox{16.2cm}{!} 
{ 
\centering
\renewcommand{\arraystretch}{1} 
\begin{tabular}[c]{cccccccccccccc}
\hline
                                            &                                                         & \multicolumn{12}{c}{\textbf{Text} (\textit{Caption})}                                                                          \\ \cline{3-14} 

                                            &                                                         & \multicolumn{4}{c|}{\textbf{-w/C (K=5)}}                                                                                                           & \multicolumn{4}{c|}{\textbf{-w/C (K=10)}}                                                                                                                              & \multicolumn{4}{c}{\textbf{-w/C (K=15)}}                                                                                      \\ \hline
\textbf{Resources}                               & \textbf{Languages}                                      & \textbf{\ft}                   & \textbf{\ff}                   & \textbf{\fm}                   & \multicolumn{1}{c|}{\textbf{\fo}}                   & \textbf{\ft}                                       & \textbf{\ff}                   & \textbf{\fm}                   & \multicolumn{1}{c|}{\textbf{\fo}}                   & \textbf{\ft}                   & \textbf{\ff}                   & \textbf{\fm}                   & \textbf{\fo}                   \\ \hline
\multicolumn{1}{c|}{}                       & \multicolumn{1}{c|}{\cellcolor[HTML]{E8E8E8}English}    & \cellcolor[HTML]{E8E8E8}50.07 & \cellcolor[HTML]{E8E8E8}43.08 & \cellcolor[HTML]{E8E8E8}52.37 & \multicolumn{1}{c|}{\cellcolor[HTML]{E8E8E8}54.61} & \cellcolor[HTML]{E8E8E8}50.74                     & \cellcolor[HTML]{E8E8E8}43.79 & \cellcolor[HTML]{E8E8E8}53.32 & \multicolumn{1}{c|}{\cellcolor[HTML]{E8E8E8}55.83} & \cellcolor[HTML]{E8E8E8}51.05 & \cellcolor[HTML]{E8E8E8}44.22 & \cellcolor[HTML]{E8E8E8}53.51 & \cellcolor[HTML]{E8E8E8}56.28 \\
\hline
\end{tabular}
}
\caption{Tags Words Evaluation : \textbf{Flan-T5}. \textbf{Selected Content} (Only Important Sentences).}
\label{table:language-tags-flant5}
\end{table*}
\begin{table*}[ht]
\resizebox{16.2cm}{!} 
{ 
\centering
\renewcommand{\arraystretch}{1} 
\begin{tabular}[c]{cccccccccccccc}
\hline
                                            &                                                         & \multicolumn{12}{c}{\textbf{Text} (\textit{Caption})}                                                                                             \\ \cline{3-14} 

                                            &                                                         & \multicolumn{4}{c|}{\textbf{-w/C (K=5)}}                                                                                                           & \multicolumn{4}{c|}{\textbf{-w/C (K=10)}}                                                                                                          & \multicolumn{4}{c}{\textbf{-w/C (K=15)}}                                                                                      \\ \hline
\textbf{Resources}                               & \textbf{Languages}                                      & \textbf{R2}                   & \textbf{RL}                   & \textbf{BLEU}                 & \multicolumn{1}{c|}{\textbf{METEOR}}                 & \textbf{R2}                   & \textbf{RL}                   & \textbf{BLEU}                 & \multicolumn{1}{c|}{\textbf{METEOR}}                 & \textbf{R2}                   & \textbf{RL}                   & \textbf{BLEU}                 & \textbf{METEOR}                 \\ \hline
\multicolumn{1}{c|}{}                       & \multicolumn{1}{c|}{\cellcolor[HTML]{E8E8E8}English}    & \cellcolor[HTML]{E8E8E8}17.39 & \cellcolor[HTML]{E8E8E8}37.47 & \cellcolor[HTML]{E8E8E8}13.76 & \multicolumn{1}{c|}{\cellcolor[HTML]{E8E8E8}31.32} & \cellcolor[HTML]{E8E8E8}17.36 & \cellcolor[HTML]{E8E8E8}37.49 & \cellcolor[HTML]{E8E8E8}13.84 & \multicolumn{1}{c|}{\cellcolor[HTML]{E8E8E8}31.44} & \cellcolor[HTML]{E8E8E8}17.37 & \cellcolor[HTML]{E8E8E8}37.53 & \cellcolor[HTML]{E8E8E8}13.87 & \cellcolor[HTML]{E8E8E8}31.51 \\
\hline
\end{tabular}
}
\caption{Headline Generation Evaluation : \textbf{mt5}. \textbf{Selected Content} (Important Sentences + Article).}
\label{table:language-headline-mt5-ia}
\end{table*}
\begin{table*}[ht]
\resizebox{16.2cm}{!} 
{ 
\centering
\renewcommand{\arraystretch}{1} 
\begin{tabular}{cccccccccccccc}
\hline
                                            &                                                         & \multicolumn{12}{c}{\textbf{Text} (\textit{Caption})}                                                     \\ \cline{3-14} 
                                            &                                                         & \multicolumn{4}{c|}{\textbf{-w/C (K=5)}}                                                                                                           & \multicolumn{4}{c|}{\textbf{-w/C (K=10)}}                                                                                                          & \multicolumn{4}{c}{\textbf{-w/C (K=15)}}                                                                                      \\ \hline
\textbf{Resources}                           & \textbf{Languages}                                      & \textbf{R2}                   & \textbf{RL}                   & \textbf{BLEU}                 & \multicolumn{1}{c|}{\textbf{METEOR}}                 & \textbf{R2}                   & \textbf{RL}                   & \textbf{BLEU}                 & \multicolumn{1}{c|}{\textbf{METEOR}}                 & \textbf{R2}                   & \textbf{RL}                   & \textbf{BLEU}                 & \textbf{METEOR}                 \\ \hline
\multicolumn{1}{c|}{}                       & \multicolumn{1}{c|}{\cellcolor[HTML]{E8E8E8}English}    & \cellcolor[HTML]{E8E8E8}17.22 & \cellcolor[HTML]{E8E8E8}37    & \cellcolor[HTML]{E8E8E8}37.31 & \multicolumn{1}{c|}{\cellcolor[HTML]{E8E8E8}31.51} & \cellcolor[HTML]{E8E8E8}17.56 & \cellcolor[HTML]{E8E8E8}37.54 & \cellcolor[HTML]{E8E8E8}14.12 & \multicolumn{1}{c|}{\cellcolor[HTML]{E8E8E8}31.85} & \cellcolor[HTML]{E8E8E8}17.58 & \cellcolor[HTML]{E8E8E8}37.52 & \cellcolor[HTML]{E8E8E8}14.19 & \cellcolor[HTML]{E8E8E8}31.85 \\
\hline
\end{tabular}
}
\caption{Headline Generation Evaluation : \textbf{mT0}. \textbf{Selected Content} (Important Sentences + Article).}
\label{table:language-headline-mt0-ia}
\end{table*}
\begin{table*}[ht]
\resizebox{16.2cm}{!} 
{ 
\centering
\renewcommand{\arraystretch}{1} 
\begin{tabular}[c]{cccccccccccccc}
\hline
                                            &                                                         & \multicolumn{12}{c}{\textbf{Text} (\textit{Caption})}                             \\ \cline{3-14} 

                                            &                                                         & \multicolumn{4}{c|}{\textbf{-w/C (K=5)}}                                                                                                           & \multicolumn{4}{c|}{\textbf{-w/C (K=10)}}                                                                                                          & \multicolumn{4}{c}{\textbf{-w/C (K=15)}}                                                                                      \\ \hline
\textbf{Resources}                               & \textbf{Languages}                                      & \textbf{R2}                   & \textbf{RL}                   & \textbf{BLEU}                 & \multicolumn{1}{c|}{\textbf{METEOR}}                 & \textbf{R2}                   & \textbf{RL}                   & \textbf{BLEU}                 & \multicolumn{1}{c|}{\textbf{METEOR}}                 & \textbf{R2}                   & \textbf{RL}                   & \textbf{BLEU}                 & \textbf{METEOR}                 \\ \hline
\multicolumn{1}{c|}{}                       & \multicolumn{1}{c|}{\cellcolor[HTML]{E8E8E8}English}    & \cellcolor[HTML]{E8E8E8}21.3  & \cellcolor[HTML]{E8E8E8}41.56 & \cellcolor[HTML]{E8E8E8}18.05 & \multicolumn{1}{c|}{\cellcolor[HTML]{E8E8E8}36.55} & \cellcolor[HTML]{E8E8E8}21.44 & \cellcolor[HTML]{E8E8E8}41.65 & \cellcolor[HTML]{E8E8E8}18.21 & \multicolumn{1}{c|}{\cellcolor[HTML]{E8E8E8}36.85} & \cellcolor[HTML]{E8E8E8}21.69 & \cellcolor[HTML]{E8E8E8}41.85 & \cellcolor[HTML]{E8E8E8}18.43 & \cellcolor[HTML]{E8E8E8}37.09 \\
\hline
\end{tabular}
}
\caption{Headline Generation Evaluation : \textbf{Flan-T5}. \textbf{Selected Content} (Important Sentences + Article)}
\label{table:language-headline-flant5-ia}
\end{table*}
\begin{table*}[ht]
\resizebox{16.2cm}{!} 
{ 
\centering
\renewcommand{\arraystretch}{1} 
\begin{tabular}{cccccccccccccc}
\hline
                                            &                                                         & \multicolumn{12}{c}{\textbf{Text} (\textit{Caption})}                                                                     \\ \cline{3-14} 

                                            &                                                         & \multicolumn{4}{c|}{\textbf{-w/C (K=5)}}                                                                                                           & \multicolumn{4}{c|}{\textbf{-w/C (K=10)}}                                                                                                          & \multicolumn{4}{c}{\textbf{-w/C (K=15)}}                                                                                      \\ \hline
\textbf{Resources}                               & \textbf{Languages}                                      & \textbf{\ft}                   & \textbf{\ff}                   & \textbf{\fm}                   & \multicolumn{1}{c|}{\textbf{\fo}}                   & \textbf{\ft}                   & \textbf{\ff}                   & \textbf{\fm}                   & \multicolumn{1}{c|}{\textbf{\fo}}                   & \textbf{\ft}                   & \textbf{\ff}                   & \textbf{\fm}                   & \textbf{\fo}                   \\ \hline
\multicolumn{1}{c|}{}                       & \multicolumn{1}{c|}{\cellcolor[HTML]{E8E8E8}English}    & \cellcolor[HTML]{E8E8E8}48.37 & \cellcolor[HTML]{E8E8E8}41.46 & \cellcolor[HTML]{E8E8E8}49.82 & \multicolumn{1}{c|}{\cellcolor[HTML]{E8E8E8}52.35} & \cellcolor[HTML]{E8E8E8}48.36 & \cellcolor[HTML]{E8E8E8}41.47 & \cellcolor[HTML]{E8E8E8}50.16 & \multicolumn{1}{c|}{\cellcolor[HTML]{E8E8E8}52.49} & \cellcolor[HTML]{E8E8E8}48.45 & \cellcolor[HTML]{E8E8E8}41.41 & \cellcolor[HTML]{E8E8E8}49.79 & \cellcolor[HTML]{E8E8E8}52.52 \\
\hline
\end{tabular}
}
\caption{Tags Words Evaluation : \textbf{mT5}. \textbf{Selected Content} (Important Sentences + Article)}
\label{table:language-tags-mt5-ia}
\end{table*}
\begin{table*}[ht]
\resizebox{16.2cm}{!} 
{ 
\centering
\renewcommand{\arraystretch}{1} 
\begin{tabular}{cccccccccccccc}
\hline
                                            &                                                         & \multicolumn{12}{c}{\textbf{Text} (\textit{Caption})}                                                                         \\ \cline{3-14} 

                                            &                                                         & \multicolumn{4}{c|}{\textbf{-w/C (K=5)}}                                                                                                           & \multicolumn{4}{c|}{\textbf{-w/C (K=10)}}                                                                                                          & \multicolumn{4}{c}{\textbf{-w/C (K=15)}}                                                                                      \\ \hline
\textbf{Resources}                               & \textbf{Languages}                                      & \textbf{\ft}                   & \textbf{\ff}                   & \textbf{\fm}                   & \multicolumn{1}{c|}{\textbf{\fo}}                   & \textbf{\ft}                   & \textbf{\ff}                   & \textbf{\fm}                   & \multicolumn{1}{c|}{\textbf{\fo}}                   & \textbf{\ft}                   & \textbf{\ff}                   & \textbf{\fm}                   & \textbf{\fo}                   \\ \hline
\multicolumn{1}{c|}{}                       & \multicolumn{1}{c|}{\cellcolor[HTML]{E8E8E8}English}    & \cellcolor[HTML]{E8E8E8}48.27 & \cellcolor[HTML]{E8E8E8}41.36 & \cellcolor[HTML]{E8E8E8}49.74 & \multicolumn{1}{c|}{\cellcolor[HTML]{E8E8E8}52.04} & \cellcolor[HTML]{E8E8E8}48.28 & \cellcolor[HTML]{E8E8E8}41.48 & \cellcolor[HTML]{E8E8E8}49.39 & \multicolumn{1}{c|}{\cellcolor[HTML]{E8E8E8}51.99} & \cellcolor[HTML]{E8E8E8}48.23 & \cellcolor[HTML]{E8E8E8}41.51 & \cellcolor[HTML]{E8E8E8}49.6  & \cellcolor[HTML]{E8E8E8}52.23 \\
\hline
\end{tabular}
}
\caption{Tags Words Evaluation : \textbf{mT0}. \textbf{Selected Content} (Important Sentences + Article)}
\label{table:language-tags-mt0-ia}
\end{table*}
\begin{table*}[ht]
\resizebox{16.2cm}{!} 
{ 
\centering
\renewcommand{\arraystretch}{1} 
\begin{tabular}[c]{cccccccccccccc}
\hline
                                            &                                                         & \multicolumn{12}{c}{\textbf{Text} (\textit{Caption})}                                                                      \\ \cline{3-14} 

                                            &                                                         & \multicolumn{4}{c|}{\textbf{-w/C (K=5)}}                                                                                                           & \multicolumn{4}{c|}{\textbf{-w/C (K=10)}}                                                                                                                              & \multicolumn{4}{c}{\textbf{-w/C (K=15)}}                                                                                      \\ \hline
\textbf{Resources}                               & \textbf{Languages}                                      & \textbf{\ft}                   & \textbf{\ff}                   & \textbf{\fm}                   & \multicolumn{1}{c|}{\textbf{\fo}}                   & \textbf{\ft}                                       & \textbf{\ff}                   & \textbf{\fm}                   & \multicolumn{1}{c|}{\textbf{\fo}}                   & \textbf{\ft}                   & \textbf{\ff}                   & \textbf{\fm}                   & \textbf{\fo}                   \\ \hline
\multicolumn{1}{c|}{}                       & \multicolumn{1}{c|}{\cellcolor[HTML]{E8E8E8}English}    & \cellcolor[HTML]{E8E8E8}50.91 & \cellcolor[HTML]{E8E8E8}44.03 & \cellcolor[HTML]{E8E8E8}53.64 & \multicolumn{1}{c|}{\cellcolor[HTML]{E8E8E8}56.42} & \cellcolor[HTML]{E8E8E8}50.92                     & \cellcolor[HTML]{E8E8E8}44.08 & \cellcolor[HTML]{E8E8E8}53.9  & \multicolumn{1}{c|}{\cellcolor[HTML]{E8E8E8}56.23} & \cellcolor[HTML]{E8E8E8}50.86 & \cellcolor[HTML]{E8E8E8}44.28 & \cellcolor[HTML]{E8E8E8}53.93 & \cellcolor[HTML]{E8E8E8}56.47 \\
\hline
\end{tabular}
}
\caption{Tags Words Evaluation : \textbf{Flan-T5}. \textbf{Selected Content} (Important Sentences + Article)}
\label{table:language-tags-flant5-ia}
\end{table*}

\end{document}